\documentclass[journal]{IEEEtran}
\usepackage[pdftex]{graphicx}
\usepackage{amsmath,array,url}
\usepackage{algorithmic,algorithm}
\usepackage{siunitx}
\usepackage{multirow}
\usepackage{tabularx, makecell, boldline}
\usepackage{hyperref}
\usepackage{cleveref}
\usepackage{subcaption}
\usepackage{xcolor}
\usepackage{color, colortbl}
\usepackage{tikz}
\usepackage{collcell}

\definecolor{Gray}{gray}{0.9}
\definecolor{LightCyan}{rgb}{0.88,1,1}
\definecolor{DarkCyan}{rgb}{0,1,1}
\definecolor{Cyan}{rgb}{0.4,1,1}
\newcolumntype{g}{>{\columncolor{Gray}}c}

\newcommand{\clsName}{DFE-NET }

\newcommand{\etal}{\textit{et al.}}

\begin{document}

\title{Surface Defect Detection and Evaluation for\break Marine Vessels using Multi-Stage Deep Learning\thanks{This work was supported primarily by the PPG Industries, Inc. 
Computation by L. Yu and J. Z. Wang has been supported in part by the Extreme Science and Engineering Discovery Environment (XSEDE), which is supported by National Science Foundation grant number ACI-1548562.}}
\author{Li Yu*, Kareem Metwaly*, James Z. Wang, and Vishal Monga\thanks{* L. Yu and K. Metwaly have equal contributions.}
\thanks{L. Yu and J. Z. Wang are with the College of Information Sciences and Technology, The Pennsylvania State University, University Park, PA 16802 USA (e-mails:
luy133@psu.edu, jwang@ist.psu.edu).}
\thanks{K. Metwaly and V. Monga are with the Department of Electrical Engineering, The Pennsylvania State University, University Park, PA 16802 USA (e-mails: kareem@psu.edu, vum4@psu.edu).}}
% make the title area
\maketitle

\begin{abstract}
Detecting and evaluating surface coating defects is important for marine vessel maintenance. 
% Trained inspectors often give inconsistent estimates of the area percentage for the detects because it is difficult to measure with bare eyes. 
Currently the assessment is carried out manually by qualified inspectors using international standards and their own experience.
Automating the processes is highly challenging because of the high level of variation in vessel type, paint surface, coatings, lighting condition, weather condition, 
paint colors, areas of the vessel, and time in service.
We present a novel deep learning-based pipeline to detect and evaluate the percentage of corrosion, fouling, and delamination on the vessel surface from normal photographs. We propose a multi-stage image processing framework, including ship section segmentation, defect segmentation, and defect classification, to automatically recognize different types of defects and measure the coverage percentage on the ship surface. 
% Experimental results demonstrate that our proposed pipeline outperforms baseline models and trained human inspectors.
Experimental results demonstrate that our proposed pipeline can objectively perform a similar assessment as a qualified inspector.
\end{abstract}

\begin{IEEEkeywords}
Defect detection, marine vessel, deep learning, image segmentation, multi-label classification.
\end{IEEEkeywords}

\IEEEpeerreviewmaketitle

\section{Introduction}
\label{intro}
Detecting and evaluating surface coating defects are important procedures in the maintenance and repair of marine vessels. Defects, such as corrosion and fouling, have to be blasted before repainting so that it maintains the hull integrity and assures the surface poses little resistance to the water. Visual inspection of defects is typically carried out at shipyards by trained human inspectors. This procedure, however, can be costly and subjective. The quality of defect assessment depends largely on the experience and attentiveness of the inspectors. An automatic defect detection system is thus highly desired to achieve comparable accuracy as humans but with higher efficiency and consistency.

The problem of automatic detection and evaluation of vessel surface defects, however, is highly challenging because of the high level of variation in vessel type, paint surface, coatings,
lighting condition, weather condition, 
paint colors, areas of the vessel, and time in service. The same set of algorithms is expected to provide accurate results for all types of vessels, under any reasonable photographic imaging condition, for any exterior surface area photographed, and regardless of the types and colors of the coating layers used on the vessel and time the existing coatings have been in service. It is unlikely one can develop a conventional algorithm that can have such a high level of robustness. Thus, we take a data-driven approach using deep learning.

In this work, we propose a multi-stage deep learning framework to automatically detect three types of defects--corrosion, delamination, and fouling--from images of marine vessels and estimate the percentage of defect coverage over different sections of the marine vessel. To train the networks, we also manually annotated hundreds of images and built a complete dataset for the analysis of different stages of the framework. To our knowledge, this is the first effort to build a real-world dataset of this scale and comprehensiveness and to develop a fully automatic defect detection and evaluation system that includes ship segmentation, ship section segmentation, defect detection, and defect classification.

\subsection{Related Work}
Over the past years, researchers have tried to incorporate computer vision techniques into the assessment of surface defects of marine vessels. Navarro \etal~\cite{navarro2010sensor} introduced a sensor system that estimates the gray-scale histogram of background and determines defect pixels by a threshold. Aijazi \etal~\cite{aijazi2016detecting} explored the HSV color space to separate the illumination invariant color component from the intensity so that histogram-based distributions and thresholds can be more effective to help spot corrosion of different shapes and sizes. Jalalian \etal~\cite{jalalian2018automatic} focused on the hue channel, instead of the complete HSV, that they utilized the Gaussian mixture model to estimate the circular hue histograms and the probability distribution of local entropy values~\cite{susan2017automatic} to identify segments of defects. Texture features are also used in the detection of defects. In~\cite{fernandez2013automated}, wavelet transform was adopted to reconstruct the image representation in terms of the frequency of components. High-frequency components, which usually happen at the coarse appearance, can be separated from low-frequency components, which correspond to flat and fine appearance. Bonn{\'\i}n-Pascual \etal~\cite{bonnin2010detection} combined color histograms and morphological properties to detect corrosion and cracks, in which they exploited a percolation model~\cite{yamaguchi2010fast} with a region-growing procedure to identify dark, narrow, and elongated sets of connected pixels. 

Though effective, these earlier  methods fall short in the following three aspects. First, the classical computer vision techniques only consider shallow color or texture features. They are not robust to changes in size, shape, illumination, etc. The existence of large, continuous defects can hardly be detected by these local-feature-based methods. Second, the reliance on handcrafted features makes the algorithms difficult to tune the parameters and learn from large amounts of labeled data. Deep neural networks, on the other hand, have many more parameters to be tuned than traditional methods, giving more capability and flexibility to capture more complex features. Third, previous work only focused on the detection of rust and corrosion, whereas other defects on the surface of marine vessels, such as delamination and fouling, are also important in vessel maintenance. 

The rise of deep learning has yielded a new generation of image segmentation algorithms with remarkable performance improvements. FCN~\cite{long2015fully} is one of the first fully convolutional networks to generate segmentation maps for images. Chen \etal~\cite{chen2014semantic} extended the FCN by adding the conditional random field (CRF) to help localize segment boundaries and achieved higher accuracy than pure FCNs. SegNet~\cite{badrinarayanan2017segnet} builds on an encoder-decoder architecture to include an encoder for feature extraction and a decoder with upsampling layers for pixel-wise classification. Inspired by FCNs and encoder-decoder models, Ronneberger \etal~\cite{ronneberger2015u} proposed the well-known UNet for medical image segmentation, which has multiple variations such as VNet~\cite{milletari2016v} and UNet++~\cite{zhou2018unet++}. Mask RCNN~\cite{he2017mask} is another class of segmentation models that uses a region proposal network (RPN) to propose bounding box candidates and extract the region of interest (RoI). It also has multiple extensions, including PANet~\cite{liu2018path}, MaskLab~\cite{chen2018masklab}, and CenterMask~\cite{lee2020centermask}. The DeepLab family~\cite{chen2017deeplab,chen2017rethinking} approaches differently by adopting the dilated convolution~\cite{yu2015multi} (a.k.a. ``atrous'' convolution) to address the decreasing resolution problem along with the network layers and the Atrous Spatial Pyramid Pooling (ASPP) to capture objects at multiple scales. 

Deep learning has been successfully applied to the field of defect or anomaly detection, such as the defects in pantograph slides~\cite{wei2019defect} and road damage~\cite{fan2019road}.
For the detection of surface defects of marine vessels, we adopt the UNet as the base model because our task is more similar to the tasks in medical image segmentation where objects don't have fixed size, shape, or boundary. In typical computer vision applications, objects carry clear semantic meanings. Take the working datasets of DeepLab~\cite{chen2017deeplab} as an example, the PASCAL VOC 2012~\cite{everingham2015pascal} and Cityscapes~\cite{cordts2016cityscapes} contain mostly well-defined object classes such as car and person. On the other hand, the coating defects of marine vessels often have intertwined and ambiguous boundaries between different defect types or with the background.

\subsection{Contributions}
We summarize our contributions as follows. \begin{itemize}
\item We constructed a first-of-its-kind complete dataset for defect detection and evaluation of the surface of marine vessels, from images. It contains 730 images with pixel-level defect annotations and 350 images with ship section annotations.
\item We proposed a fully automatic evaluation system to detect defects and calculate defect coverage. The system consists of four deep learning modules: whole ship segmentation, ship section segmentation, defect segmentation, and defect classification. 
\end{itemize}

\begin{figure}[ht!]
    \centering
    \includegraphics[width=0.48\textwidth]{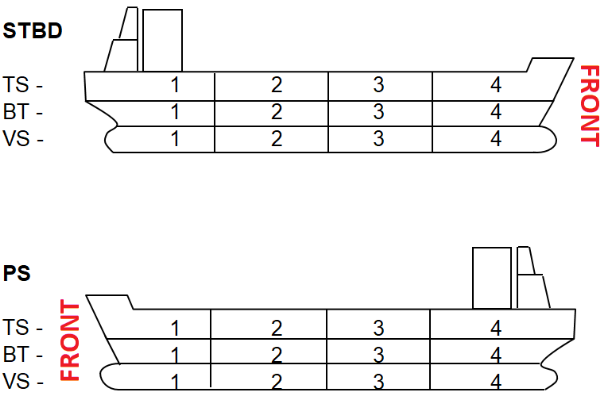}
    \caption{Sections of a ship viewed from the starboard (STBD) and the port side (PS). The three levels we used are Top Side (TS), Boot Top (BT), and Vertical Side (VS). Diagrams courtesy of the PPG Industries, Inc.}
    \label{fig:ship}
 \end{figure}

The novel aspects of the individual modules are listed below.
\begin{itemize}
    \item We proposed to segment three horizontal sections, namely Top Side (TS), Boot Top (BT), and Vertical Side (VS) (Figure~\ref{fig:ship}), of a ship by predicting the two boundaries, which converts 2D segment prediction into 1D curve prediction. It reduced the requirement for labeled data and increased consistency in block predictions.
    \item We utilized a teacher-student training scheme in defect segmentation to overcome the shortage of labeled data and the problem of coarse or incomplete labels. The teacher model was first trained on manually labeled defect data and used to generate pseudo labels. The pseudo labels were then combined with the original labels to form more accurate and complete labels. The student model trained on the new labels outperformed the teacher model.
    \item We adopted spatial transformer networks (STN)~\cite{jaderberg2015spatial} as feature extractors and proposed a multi-head architecture for multi-label prediction. STNs can effectively estimate affine transformations such that input image patches are warped for better feature extraction. A delamination detection head was added aside to the general head to cope with the need to detect the more challenging delamination defects.
\end{itemize}

The remainder of the paper is organized as follows. Section~\ref{sec:dataset} describes the details of different defects and our collected datasets. Section~\ref{sec:approach} sketches the multi-stage deep learning pipeline and the subsequent Sections~\ref{sec:partseg} and~\ref{sec:defectsegcls} present the architectures of individual modules. Experiment results, including ablation studies and comparisons with baseline models, are reported in Section~\ref{sec:expresult} . Finally, we conclude and suggest future work in Section~\ref{sec:conclusion}.

\section{The Dataset}
\label{sec:dataset}
\begin{figure}[ht!]
    \centering
    \includegraphics[width=0.48\textwidth]{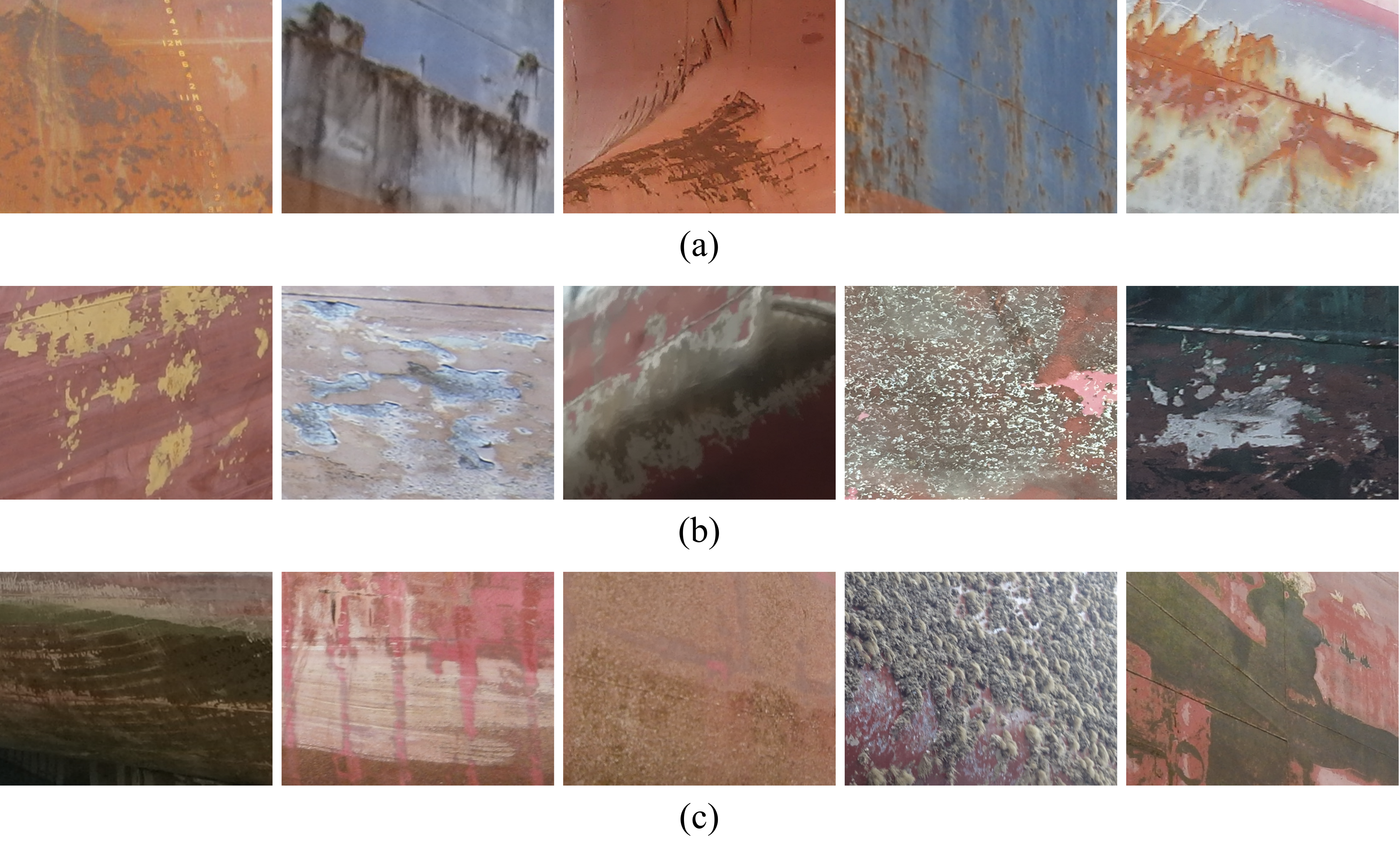}
    \caption{Typical defects of (a) corrosion, (b) delamination, and (c) fouling.}
    \label{fig:defect_illus}
 \end{figure}

\begin{figure}[ht!]
    \centering
    \includegraphics[width=0.4\textwidth]{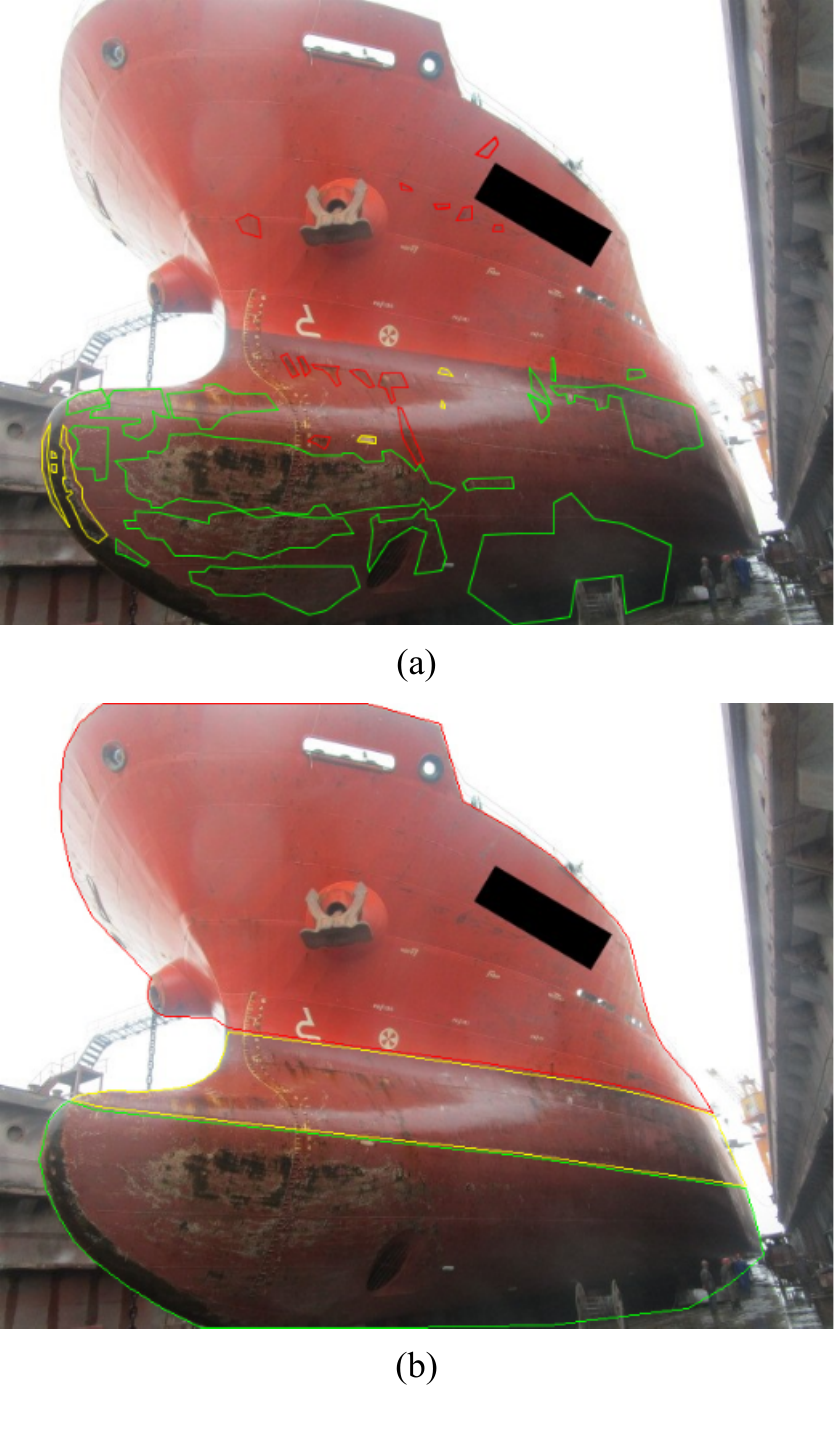}
    \caption{Example defect labels and ship section labels: (a) corrosion (red), delamination (yellow), and fouling (green); (b) TS (red), BT (yellow), and VS (green). The ship name is blocked in black for preserving confidentiality.}
    \label{fig:ship_illus}
 \end{figure}

We collaborated with the PPG Industries, Inc. (PPG) and collected a fully annotated real-world dataset. Figure~\ref{fig:defect_illus} shows some typical examples of corrosion, delamination, and fouling. They may occur in different sections of a marine vessel, as illustrated in Figure~\ref{fig:ship_illus}. Figure~\ref{fig:ship_illus}(a) displays the labels annotated by our collaborators from PPG, with different colors indicating different types of defects. To calculate the percentage of defect coverage over different sections of the ship, we need to identify sections of the ship from an image. Figure~\ref{fig:ship_illus}(b) shows the labels of three different sections. From top to bottom, they are TS, BT, and VS sections. 
We labeled around 730 images for defect detection and 350 images for section segmentation. Because the annotators have to manually classify defects and carefully trace the boundaries, the process demands expertise in marine coating and is time-consuming.

These photographs were collected by field inspectors over the years as they were doing on-site vessel inspections. Because the images were taken {\it not} for the purpose of computer-based analysis, there was no set standard on photo-taking. This poses a serious challenge to computer-based analyses because the developed algorithms must be robust enough to handle any reasonable in-the-wild photo-taking condition.

% \subsection{Defect Types and Measurements}
 
% \subsection{Dataset}

\section{Multi-Stage Approach Overview}
\label{sec:approach}
\begin{figure}
    \centering
    \includegraphics[width=0.48\textwidth]{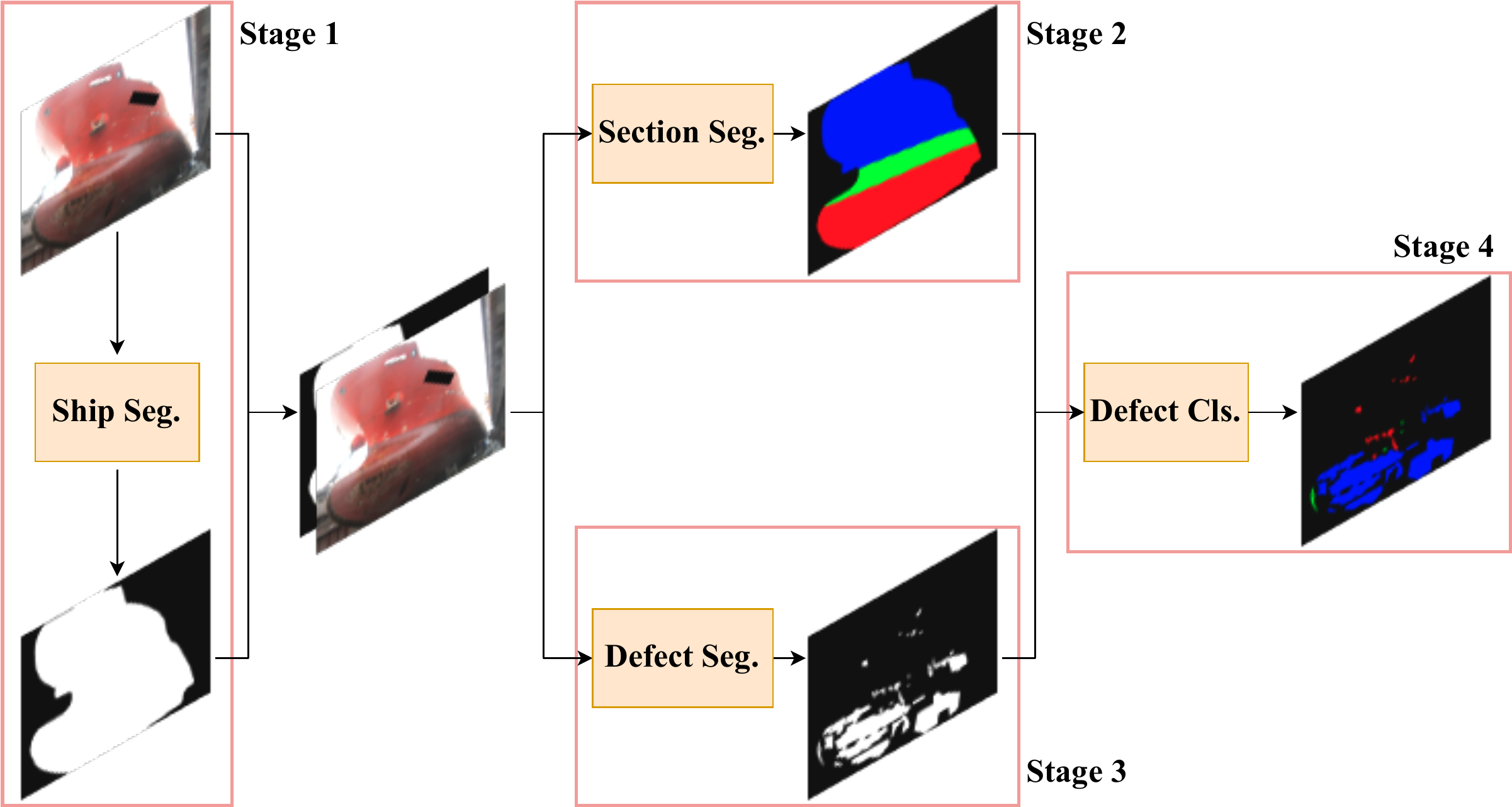}
    \caption{The multi-stage pipeline used in our system.}
    \label{fig:pipeline}
 \end{figure}

The multi-stage approach for defect detection mainly consists of four stages, as shown in Figure~\ref{fig:pipeline}. They are whole ship detection, section segmentation, defect segmentation, and classification. Stages 1 and 2 are required for extracting the region corresponding to the ship and subsequently  
 different sections of the ship. Stages 3 and 4 perform defect segmentation and classification respectively. We contend that performing this as two-serial steps is superior to a single multi-class segmentation algorithm, because a region may contain multiple  defect types. That is, an object in an image can have two labels, making typical multi-class segmentation models unsuitable.

\section{Whole Ship and Section Segmentation}
\label{sec:partseg}
In order to evaluate the percentage of three defect types on different sections of the ship, we first isolate the ship within the image and segment different sections. To isolate the ship, we use an UNet~\cite{ronneberger2015u} based image segmentation algorithm. The whole ship segmentation model is trained under the assumption that only a single prominent ship exists in an image. Those less prominent or partially blocked ships will be ignored. This ensures that we focus on one ship in each image for defect detection and evaluation.

For ship section segmentation, we propose to view the three sections ({\it i.e.}, TS, BT, and VS) as separated by two curves between TS and BT, and BT and VS, so that section segmentation is transformed from predicting 2D maps to predicting 1D boundaries. An RNN-based line smoothing module is introduced to ensure consistency and smoothness of the predicted 1D line boundaries. We base our model on HorizonNet~\cite{sun2019horizonnet}, which was proposed to predict room layouts from panoramic images, and made some major changes to adapt to our scenario. The following subsections will describe the model architecture and some modifications.

\subsection{Network Architecture}
\begin{figure}
    \centering
    \includegraphics[width=0.8\linewidth]{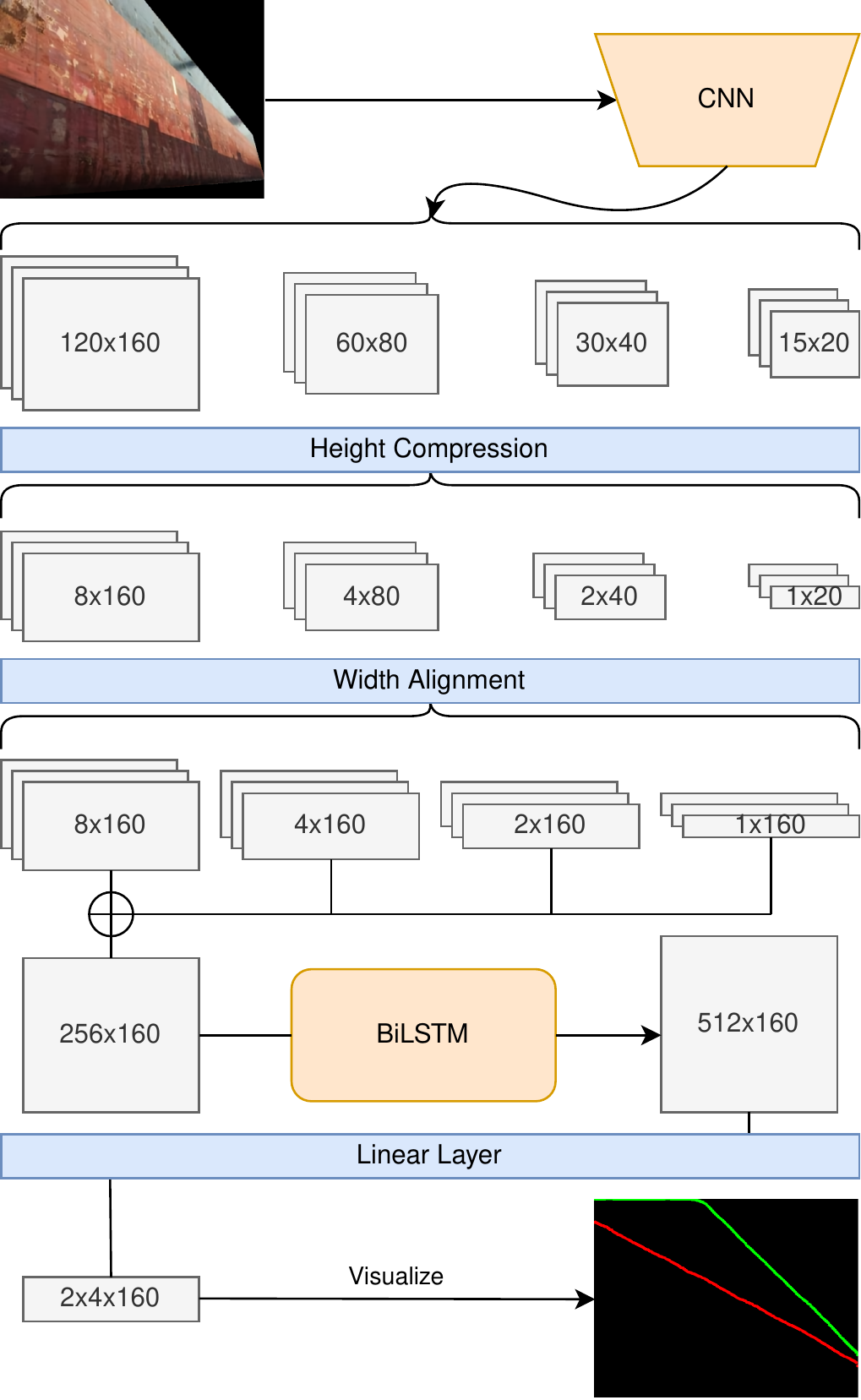}
    \caption{Architecture of the section segmentation network.}
    \label{fig:segpart_archi}
\end{figure}

The architecture for section segmentation is illustrated in Figure~\ref{fig:segpart_archi}. The input is the cropped ship area from whole ship segmentation and the output are two 1D vectors representing the two boundary curves TS/BT and BT/VS. The input image is cropped and resized to $640\times480$ so that we can have fixed-sized input and output. We use a ResNet-based~\cite{he2016deep} convolutional neural network (CNN) as a feature extractor to extract the feature maps at four different scales. A height compression module is adopted to compress the feature maps at the vertical direction, which is achieved by four consecutive down-sampling convolutions. Each convolution layer has a stride of 2 at the vertical dimension and a stride of 1 at the horizontal dimension with proper padding to make sure that the height gets compressed while the width is maintained. The resulting flat feature vectors are then aligned to the same width and stacked into a large feature map. Different from the HorizonNet which aligns the width with direct interpolation, we use a fully connected layer to convert the width of feature maps into a uniform sequence length of 160. After we get the large feature map of size $256\times160$, we can use a linear layer (fully connected layer) to generate the desired output. However, as indicated in HorizonNet, we found the immediate output is not smooth and suffers from some defects. Thus we use a Bidirectional LSTM~\cite{hochreiter1997long} to smooth the concatenated feature map, before generating the final curve lines.

\subsection{Range-Aware Loss Function}
As can be seen from Figure~\ref{fig:segpart_archi}, the boundary between TS and BT or between BT and VS does not always span the full width of 640 pixels, even though we have cropped the ship and resized it to fit into the entire input image. The missing part of the green line (big chunk on the top and small chunk on the right) should not be considered in evaluating the performance of our model. Because different ships come with different lengths or spans of boundaries, we cannot make the model accept ground truth labels with variable length while predicting the two curves with exact length. In fact, both the ground truth and predicted labels of our model are of the shape $2\times640$. To alleviate the mismatch of real boundary length and the fixed-length required by the model, we propose a range-aware loss function that applies a mask over the range of the real boundary and only calculates loss for the range during training. The loss function can be defined as:
\begin{equation}
    \mathcal{L} = \sum_{i=1}^2 \sum_{j=1}^{640} m(i,j) |y_{i,j}^t - y_{i,j}^p|\;,
\end{equation}
where $i \in \{1,2\}$ is the index for the two boundaries and $j \in [1,640]$ is the position index in the width direction.  $m(i,j) \in \{0,1\}$ is a mask function to indicate if a prediction $y_{i,j}^p$ for the $i$th boundary at the $j$th position needs to be evaluated. Note that sometimes there are only two sections or even one section in an image, which means we may have one or zero boundary as ground truth, we set $m(i,j)$ as all zeros for the missing boundaries and thus the predictions for them are ignored.

\subsection{Data Processing and Augmentation}
As mentioned in Section~\ref{sec:dataset}, the raw dataset contains both high-quality images ({\it e.g.} of size $4000\times3000$) and highly-compressed, low-quality images ({\it e.g.} of size $400\times300$), we have to preprocess the data before they can be used for training. For whole ship segmentation, we resize raw images and labeled ship masks into a uniform size $640\times480$ and use them as input. The output ship masks are of the same size. We then use the predicted ship masks to crop ship areas and resize the ships again to size $640\times480$. The cropped ships and correspondingly processed section boundary labels, as shown in Figure~\ref{fig:segpart_archi}, are finally used for ship section segmentation.

We also perform active data augmentation in section segmentation to make the best use of the 350 labeled images. Because the raw images were photos taken from various angles and distances from the ships, the ship areas presented in images can be various portions of the entire ships, from different perspectives. To enrich the diversity of training data and to make the trained model more robust, we randomly rotate and shift the ship areas within a small range and repeat it several times. Moreover, ships come in different colors, with the majority of ships in red color while others in blue color or white color. We believe the color of ships should not affect the determination of the three sections, so we also flip the RGB color channels to BGR as a special data augmentation strategy. As a result, we can generate 3,074 images for training. 
% Some example augmented images for one ship are shown in Fig.~\ref{fig:aug_example}. The ship on top left is the original one and the others are generated by random rotation, shift, color flip or all that combined.
% \begin{figure}
%     \centering
%     \includegraphics[width=\linewidth]{imgs/whole_part_seg/aug_example.pdf}
%     \caption{Some example images of data augmentation for ship part segmentation}
%     \label{fig:aug_example}
% \end{figure}

\section{Defect Segmentation and Classification}
\label{sec:defectsegcls}
\subsection{Defect Segmentation}
\label{subsec:defectseg}
\begin{figure}
    \centering
    \includegraphics[width=\linewidth]{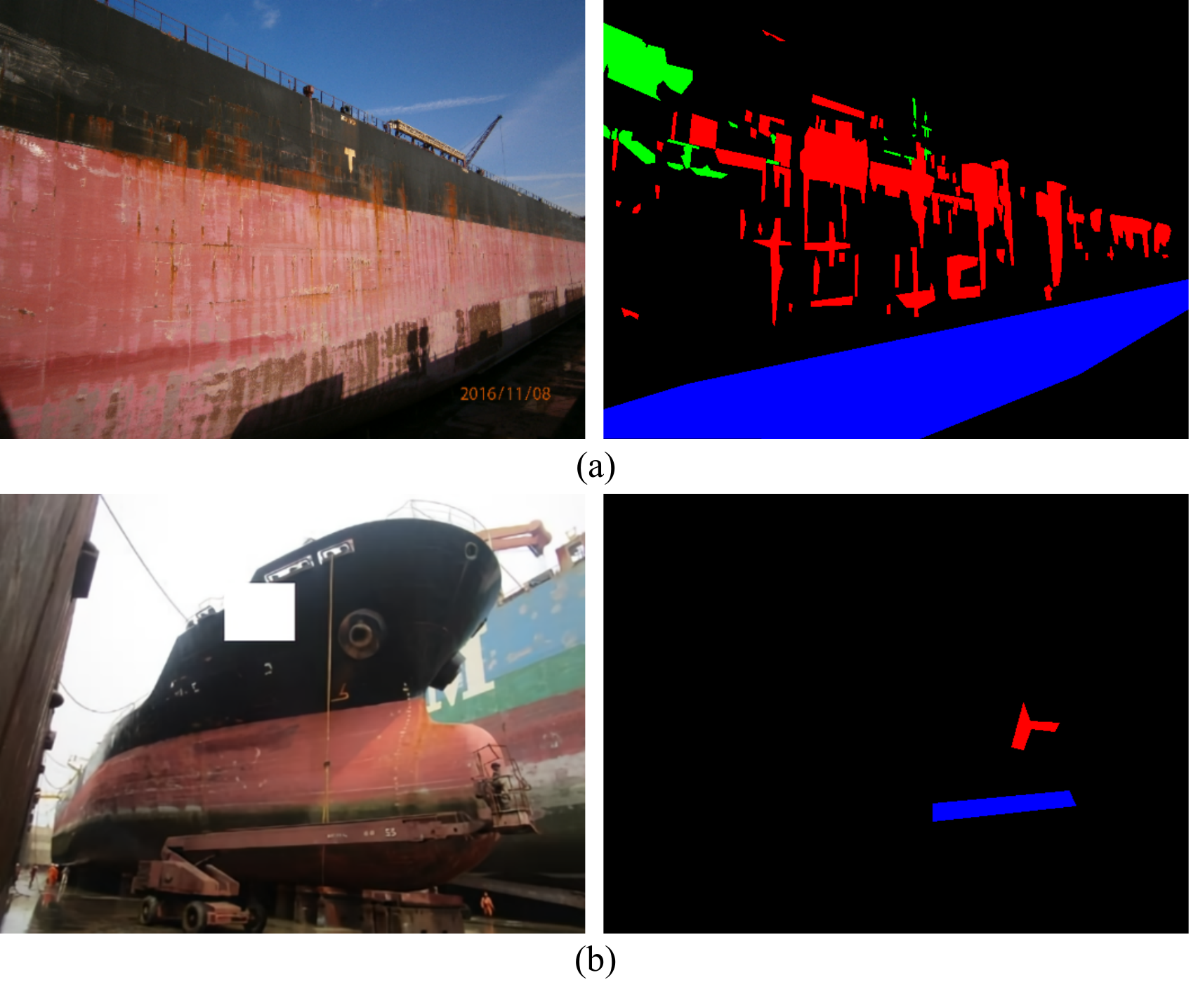}
    \caption{Examples of weak labels: polygons too coarse (a) and not complete (b). Different colors indicate different types of defects, with red for corrosion, green for delamination, and blue for fouling.}
    \label{fig:weak_label_illus}
\end{figure}

We apply weakly-supervised learning to the segmentation of defects, as we find the quality of labels is not always consistent. Some labels tend to be more ``coarse'' than others. Take the large blue polygon in Figure~\ref{fig:weak_label_illus}(a) as an example, it is supposed to mark those black areas of the ship on the left as fouling, but unnecessarily includes the pale regions which should be regarded as background. On the other hand, some areas with defects are neglected and unlabeled. The ship in Figure~\ref{fig:weak_label_illus}(b) has a broad, greenish fouling defect at the bottom, but it is only partially labeled. The blue polygon on the right covers a small part of fouling.

\begin{figure}
    \centering
    \includegraphics[width=\linewidth]{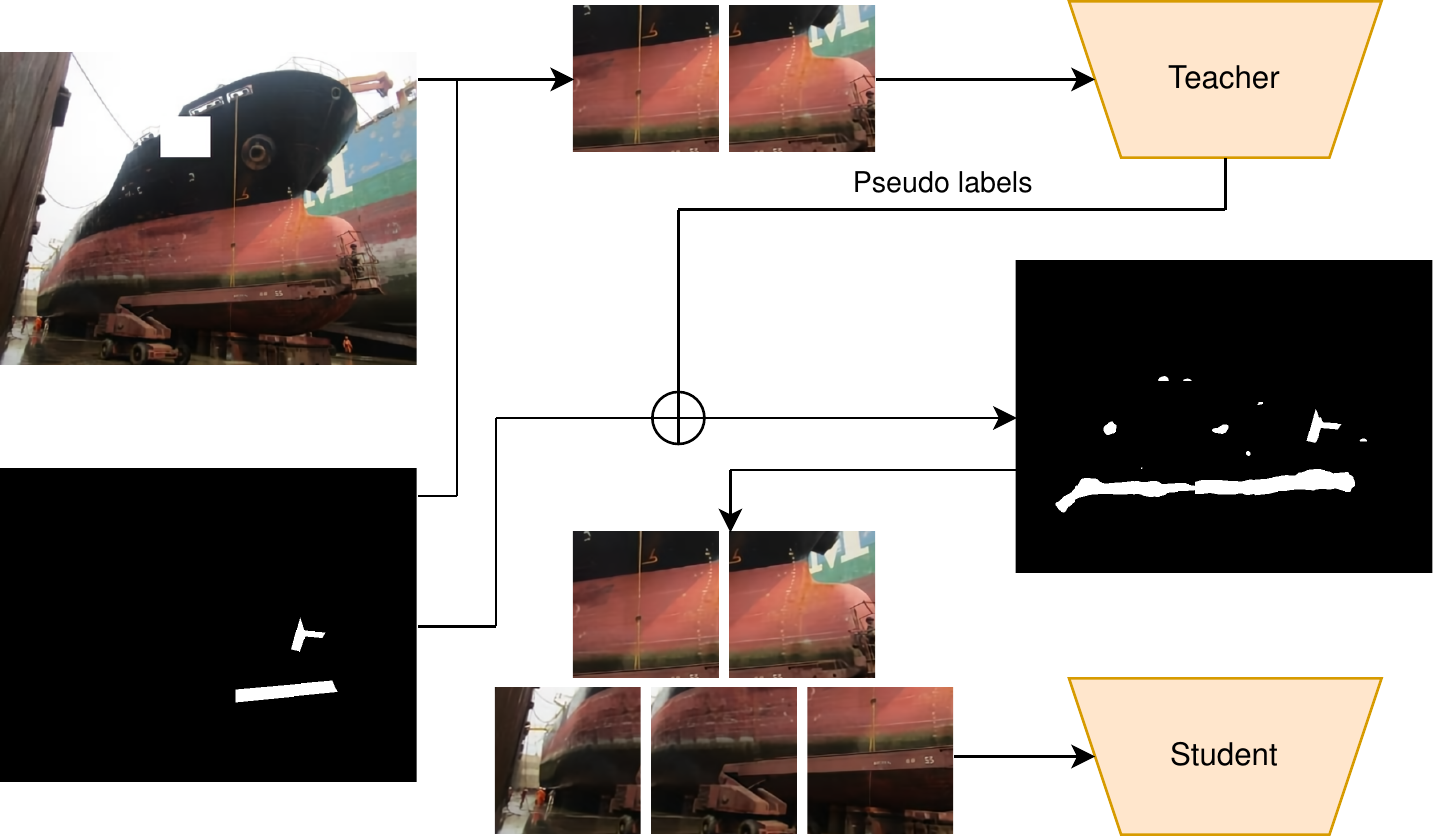}
    \caption{The teacher-student training scheme.}
    \label{fig:teastu}
\end{figure}

To mitigate the negative effects of weak annotations, we propose to utilize a teacher-student training scheme for the defect segmentation task. Similar to GrabCut~\cite{rother2004grabcut} and BoxSup~\cite{dai2015boxsup}, which start with bounding box annotations and gradually update segmentation masks via iterative model training, we train one model on source labels and use the predictions as input to train another model. As plotted in Figure~\ref{fig:teastu}, a teacher model is trained on the original label data, to generate pseudo labels. The predicted labels will get combined with the original labels and then be used to train a student model. Both teacher and student models are UNet-like image segmentation models~\cite{ronneberger2015u}, with ResNet34~\cite{he2016deep} as the backbone. We choose UNet because it is widely used in medical image segmentation tasks and the form of defects in our dataset resembles more of the objects in medical imaging rather than those in natural scenes. A shallow backbone ResNet34 is used instead of deeper feature extractors such as ResNet50 or DenseNet121~\cite{huang2017densely}, as defects are mostly local features which demands less high-level feature extraction and learning. 

For training of the teacher model, we slice image-and-label pairs into patches of size $224\times224$ and remove those without defects or with defect areas less than one percent of the patch. We observed in the labeled data and during experiments that the problem of missing labels is more prominent than coarse labels. The existence of false backgrounds (unlabeled defects) in training data would impact more on the performance than false foregrounds (background labeled as defects). This is reasonable because the areas that are viewed and labeled by human annotators are more reliable than the others that are unattended. That is, though labeled polygons may contain excess regions, they are more accurate than unlabeled regions. As a result, around half of the patches are removed. 

After the teacher model is trained for a few epochs, it is used to predict defect labels on the training images. The predicted labels are then combined with the original labels as a new set of labels to generate a new set of training data. Note that by taking the union of pseudo and human labels, we only attempt to solve the problem of missing labels. As aforementioned, it is more prominent than the problem of coarse labels. We also attempted to take the intersection of predicted and human labels at patches containing mostly fouling (coarse labels happen more in fouling than other two defects), the segmentation result, however, was worse than only using human labels. It suggests that even trained on refined data, the teacher model is not able to predict all real fouling correctly. We have to keep the coarse polygons in the source labels to ensure better training of the student model. Finally, the student is trained on updated labels and let run for full epochs to get the final model.
\subsection{Defect Classification}
\label{subsec:defectcls}
For defect classification, we use a multi-label algorithm rather than a multi-class one. It enables the detection and classification of overlapping different types of defects rather than only detecting the most prominent one as in the multi-class analogue. In other words, we answer three different Yes/No questions for different defect types, respectively, rather than answering one question with the type of defect.

Moreover, in analyzing defect categories, we have noticed that delamination is the most challenging one to detect. Therefore, we use an additional feature extractor (dubbed Delamination Feature Extractor or DFE) in the network architecture which pays more attention to features related to the detection of delamination defects. To ensure the effectiveness of the added (delamination) feature extractor, we regularize the loss function in training such that the added feature extractor is generating complementary information rather than the same features of the main (general) feature extractor.

Last but not least, we prepend the two feature extractors with spatial transformer networks (STN)~\cite{jaderberg2015spatial}. STN can be used to estimate the parameters of affine transformations that can be used to warp the input image patch before passing by the feature extractors. This effectively enables both feature extractors to have different patches as inputs, where each generated patch focuses on the most relevant region of the original input patch. Besides, it enables the network to adaptively crop the input patch to the region that only has defects, {\it i.e.} focusing on the most important region of a patch. Thus, we have a better classification of the input image patch.

\subsubsection{\textbf{Network architecture}}
\begin{figure}
    \centering
    \includegraphics[width=\linewidth]{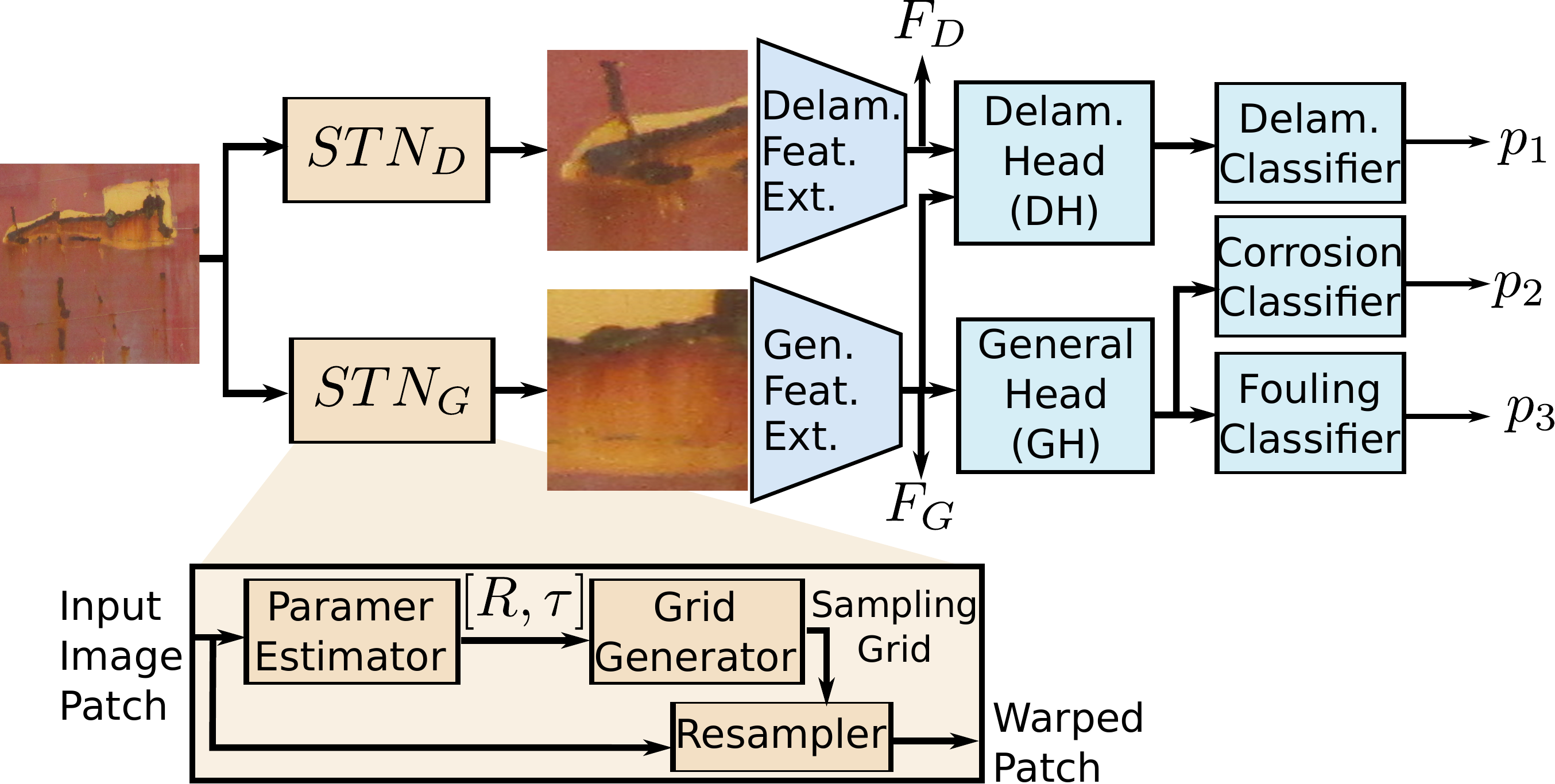}
    \caption{Architecture of the defect classification network.}
    \label{fig:cls_netw}
\end{figure}

As depicted in \Cref{fig:cls_netw}, the network takes patches of images. The preceding defect segmentation network detects regions of interest (RoI) for defects. Then, we consider patches of the input images where we have an RoI ratio $> 0.1$, {\it i.e.} we only consider patches of the input image where the defect segmentation suspects that more than $0.1$ of the area of the patch is marked as defected. This way the classification algorithm would select patches that have a reasonable area of defects rather than selecting patches arbitrarily. We empirically consider patches of size $64\times64$ as it is not too large to cause mosaic artifact in the final classification image while being not too small to allow utilization of information from surrounding pixels (such as texture and edges). 

Each input patch pass through two STN networks (more details in \Cref{subsubsec:stn}) that enable the following feature extractors to have different inputs focusing on different regions of the input image patch, if needed. The structure of the STN network is shown in \Cref{tab:stn}. The output of each STN is a warped copy of the input image patch. We use Densenet-121 \cite{huang2017densely} as the backbone for our feature extractors which is initially pre-trained over the ImageNet dataset \cite{imagenet_cvpr09}. Densenet is a deep architecture that has skip connections between each layer and all the following layers in each dense block. This ensures the flow of information from one layer to deeper layers, thus mitigating the vanishing gradient problem with deeper architectures \cite{tan2019vanishing}. We do not use deeper than Densenet-121 to save computational resources while mitigating the overfitting problem that arises with deeper architectures with limited training data \cite{metwaly2020attention}. After extracting the features, an average pool operation is used to eliminate the height and width dimensions, thus we end up with a one-dimensional vector of length $1024$. The average pool operation enables the network to work with any arbitrary size because the latter layers are fully connected. The structure of the feature extractors is detailed in \Cref{tab:cls_feat_ext}. The extracted two feature vectors are denoted by $F_D$ and $F_G$ for Delamination-specific features and general features, respectively. $F_G$ passes then by two fully-connected layers (named General Head  or GH) with a ReLU activation to reduce the feature size to $256$. Similarly, we concatenate $F_G$ and $F_D$ and pass it by Delamination Head (DH) to reduce the size to $256$. At the last stage, three fully connected networks of two layers are used as the last classification stage for each class independently. The first layer has a ReLU activation function while the last one has a Sigmoid activation. Therefore, the outputs of the sub-networks $p_1$, $p_2$ and $p_3$ are three values that range from $0$ to $1$ which represents the probabilities of different type of defects (Corrosion, Fouling or Delamination). The details of the GH, DH, and the classifier sub-networks are shown in \Cref{tab:cls_fcn}.

\begin{table*}
\centering
\caption{Structure of the Spatial Transformer Networks (STNs)}
\resizebox{\textwidth}{!}{%
\begin{tabular}{l|cccccc}\Xhline{4\arrayrulewidth}
      & STN.1  & STN.2 & STN.3 & STN.4 & GridGen & Resampler \\\hline
Input & $\begin{matrix}\text{Image Patch} \\64\times64\times3\end{matrix}$ & STN.1 &  STN.2 & STN.3 & STN.4 & [Image Patch, GridGen]\\
Structure & $\begin{bmatrix}7\times7~\text{conv.}\\\text{ReLU}\end{bmatrix}$ & $\begin{bmatrix}5\times5~\text{conv.}\\\text{ReLU}\end{bmatrix}$ & $\begin{bmatrix}3\times3~\text{conv.}\\\text{ReLU}\end{bmatrix}$ & $\begin{bmatrix}1\times1~\text{conv.}\\\text{avg-pool} \end{bmatrix}$ & $\begin{matrix}\text{Sample Grid}\\\text{Generator}\end{matrix}$& $\begin{matrix}\text{2D interpolation based}\\\text{on the sampling grid}\end{matrix}$\\
Output & $64\times64\times100$ & $64\times64\times100$ & $64\times64\times50$ & $1\times1\times6$ & $64\times64\times1$ & $64\times64\times3$ \\\hline
\end{tabular}}
\label{tab:stn}
\end{table*}

\begin{table*}
\centering
\caption{Structure of the feature extractors sub-networks -- General Feature Extractor and Delamination Feature Extractor}
\resizebox{\textwidth}{!}{%
\begin{tabular}{l|cccccccc}\Xhline{4\arrayrulewidth}
      & Base  & Dense.1 & Trans.1 & Dense.2 & Trans.2  & Dense.3  & Trans.3 & Dense.4\\\hline
Input & $\begin{matrix}\text{Warped Image Patch} \\64\times64\times3\end{matrix}$ & Base &  Dense.1 & Trans.1 & Dense.2  & Trans.2  & Dense.3 & Trans.3\\
Structure & $\begin{bmatrix}7\times7~\text{conv.}\\3\times3~\text{max-pool}\end{bmatrix}$ & $\begin{bmatrix}1\times1~\text{conv.}\\3\times3~\text{conv.} \end{bmatrix}\times6$& $\begin{bmatrix}1\times1~\text{conv.}\\2\times2~\text{avg-pool} \end{bmatrix}$&$\begin{bmatrix}1\times1~\text{conv.}\\3\times3~\text{conv.} \end{bmatrix}\times12$ & $\begin{bmatrix}1\times1~\text{conv.}\\2\times2~\text{avg-pool} \end{bmatrix}$ & $\begin{bmatrix}1\times1~\text{conv.}\\3\times3~\text{conv.}\end{bmatrix}\times24$ & $\begin{bmatrix}1\times1~\text{conv.}\\2\times2~\text{avg-pool} \end{bmatrix}$ & $\begin{bmatrix}1\times1~\text{conv.}\\3\times3~\text{conv.}\end{bmatrix}\times16$\\
Output & $16\times16\times64$ & $16\times16\times256$ & $8\times8\times128$ & $8\times8\times512$ & $4\times4\times256$ & $4\times4\times1024$ & $2\times2\times512$ & $1\times1\times1024$ \\\hline
\end{tabular}}
\label{tab:cls_feat_ext}
\end{table*}

\begin{table*}
\centering
\caption{Structure of the classifiers}
\resizebox{\textwidth}{!}{%
\begin{tabular}{l|cc|cc|cc}\Xhline{4\arrayrulewidth}
      & \multicolumn{2}{c|}{General Head} &  \multicolumn{2}{c}{Delamination Head} & \multicolumn{2}{c}{Classifiers}\\\cline{2-7}
      & GH.1  & GH.2 & DH.1 & DH.2 & C.1 & C.2\\\hline
Input & $F_G$ & GH.1 & $[F_G, F_D]$ & DH.1 & GH.2 or DH.2 & C.1\\
Structure & $\begin{bmatrix}1024\rightarrow512~\text{Linear}\\\text{ReLU}\\\text{Dropout}\end{bmatrix}$ & $\begin{bmatrix}512\rightarrow256~\text{Linear}\\\text{ReLU}\\\text{Dropout}\end{bmatrix}$ & $\begin{bmatrix}2048\rightarrow512~\text{Linear}\\\text{ReLU}\\\text{Dropout}\end{bmatrix}$ & $\begin{bmatrix}512\rightarrow256~\text{Linear}\\\text{ReLU}\\\text{Dropout}\end{bmatrix}$ & $\begin{bmatrix}256\rightarrow128~\text{Linear}\\\text{ReLU}\\\text{Dropout}\end{bmatrix}$ & $\begin{bmatrix}128\rightarrow1~\text{Linear}\\\sigma\\\end{bmatrix}$ \\
Output & $512\times1$ & $256\times1$ & $512\times1$ & $256\times1$ & $128\times1$ & $1$\\\hline
\end{tabular}}
\label{tab:cls_fcn}
\end{table*}

\subsubsection{\textbf{Delamination Feature Extractor and regularization}}
Since delamination defects are the most difficult type of defects to detect and classify, we use an additional feature extractor specifically for delamination defects. The sole purpose of that special feature extractor is to find additional features that would make it easier to detect delamination defects. However, to ensure the validity of the added feature extractor and that it is performing its expected objective, we add a regularization term to the loss function which ensures that the newly extracted features are complementary to the features extracted by the general feature extractor. Therefore, our customized training loss function is as follows.
\begin{equation}
    \mathcal{L} = \sum_{i=1}^3 w_i \mathcal{L}_\text{BCE}^{(i)}(l_i, p_i) + \lambda \cdot \left|\text{CosSim}\left(F_G, F_D\right)\right|\;,
\end{equation}
where $\mathcal{L}_\text{BCE}^{(i)}(l_i, p_i)$ is the Binary Cross Entropy \cite{goodfellow2016deep} which is defined as:
\begin{equation}
    \mathcal{L}_\text{BCE}^{(i)}(l_i, p_i) = l_i \log{p_i} + (1 - l_i) \log(1 - p_i)\;,
\end{equation}
where $l_i\in \{0,1\}$ is the ground-truth label of class $i\in \{1,2,3\}$ representing corrosion, fouling, and delamination, respectively. $p_i$ is the predicted output of class $i$ which takes values between $0$ and $1$. We multiply each BCE loss term with a weight value $w_i$ to compensate for the imbalance in the labeled dataset. We set $w_1=1$, $w_2=2$ and $w_3=4$. This way we implicitly give more focus to fouling and delamination defects in the training process over the corrosion defects, because corrosion defects are the easiest type of defects to classify in comparison to delamination and fouling defects.

The regularization term $\text{CosSim}\left(F_G, F_D\right)$ is the cosine similarity \cite{pellegrini2019cosine} between the extracted features from the general feature extractor $F_G$ and the delamination feature extractor $F_D$. $\lambda$ is a regularization parameter to tune the importance of the regularization term in the training process in comparison to the BCE loss term. We set $\lambda$ to $1$ through training by cross-validation \cite{monga_handbook_2018}. The objective of the cosine similarity regularization is to ensure that $F_D \neq \alpha \cdot F_G$ for some scale factor $\alpha$. Therefore, $F_D$ carries complementary information to $F_G$ which helps in the final classification stage. $\text{CosSim}$ measures the angle between the two latent feature vectors and attempts to maximize it by minimizing the inner product of the two latent vectors. In other words, we train our network in a weakly-supervised manner since the size of the training set is not large enough. 
$\text{CosSim}\left(x, y\right) = x^T y / \max\left(||x||\cdot ||y||, \epsilon\right)\;$. 
The inner product is divided by $\max\left(||x||\cdot ||y||, \epsilon\right)$ for normalization, where $\epsilon = 10^{-4}$ for numerical stability.

\subsubsection{\textbf{Spatial Transformer Network (STN)}}\label{subsubsec:stn}

Before passing the input image patch by the two feature extractors, we first pass it by two STNs to perform affine transformations \cite{gu2020improving} to make it easier for each feature extractors to find the most relevant features. The Spatial Transformer Network \cite{jaderberg2015spatial} estimates 6 parameters to perform an affine transformation which consists of 4 parameters to perform rotation and scaling $R$ and 2 parameters for translation $\tau$ as follows.
\begin{equation}
\begin{split}
    \left(\begin{array}{c}
         x^{(n)}_j\\
         y^{(n)}_j
    \end{array}\right) &= \mathcal{T}_\theta\left(\left(\begin{array}{c}
         x_j\\
         y_j
    \end{array}\right)\right) \\
    &= \underbrace{\left(\begin{array}{cc}
         r_{xx} & r_{xy} \\
         r_{yx} & r_{yy}
    \end{array}\right.}_{R}\underbrace{\left.\begin{array}{c}
        t_x \\
        t_y
    \end{array}\right)}_{\tau} \times \left(\begin{array}{c}
         x_j\\
         y_j\\
         1
    \end{array}\right)\;,
\end{split}
\end{equation}
where $x_j \& y_j$ are the coordinates of pixel $j$ of the original input patch and $x^{(n)}_j \& y^{(n)}_j$ are the new coordinates. 

Utilization of the STN modules in \clsName allows each feature extractor to obtain different features. It makes the problem much easier for the subsequent classifier subnetwroks. As depicted in \Cref{fig:cls_netw}, each STN warps the input image differently. For instance, delamination defects usually depend on information related to edges (as it is a result of the detachment of the paint from the surface). Therefore, warping the input image patch to see edges more prominently will make the detection process of delamination defects easier. This is performed by $STN_D$. However, corrosion and fouling defects usually depend on information related to the texture and colors (not edges unlike delamination). Therefore, $STN_G$ needs to warp the image differently to focus on the texture not the boundary changes from one surface to another. 

%We have performed several ablation studies to validate our claims in \Cref{subsec:ablation_study}.

\section{Experimental Results}
\label{sec:expresult}
\subsection{Ablation Study} \label{subsec:ablation_study}
We have performed an extensive set of ablation studies to verify and validate the importance of each module. We start by discussing the effectiveness of some modules related to the final classification network such as the additional encoder for delamination defects, the regularization term in the loss function, and the STN modules. Then, we study the effectiveness of employing the section segmentation results in the classification process as well.

\subsubsection{\textbf{Effectiveness of the Delamination Feature Extractor and regularization in the classification network}} \label{subsubsec:ablation_dfe}
To study the effectiveness of the special feature extractor, that is the Delamination Feature Extractor (DFE), we train the classification network for $20$ epochs by randomly choosing 500 images for training and 20 images for testing and repeating the same training/test split for three times then we calculate the average values of accuracy, balanced accuracy (B. Accuracy), F1-score and confusion matrices as follows.

\begin{align}
    \textsc{accuracy} &= \frac{\textsc{TP} + \textsc{TN}}{\textsc{TP} + \textsc{TN} + \textsc{FP} + \textsc{FN}}\;,\\\
    \textsc{B. Accuracy} &= \frac{1}{2}\left(\frac{\textsc{TP}}{\textsc{TP} + \textsc{FN}} + \frac{\textsc{TN}}{\textsc{TN} + \textsc{FP}}\right)\;,\\
    F_1 &= 2\frac{\textsc{Precision}\cdot\textsc{Recall}}{\textsc{Precision} + \textsc{Recall}}\;,\\
    \textsc{Precision} &= \frac{\textsc{TP}}{\textsc{TP} + \textsc{FP}}\;,\\
    \textsc{Recall} &= \frac{\textsc{TP}}{\textsc{TP} + \textsc{FN}}\;,
\end{align}
where TP, TN, FP, and FN are acronyms for True Positive, True Negative, False Positive, and False Negative, respectively.

\Cref{tab:ablation_corrosion_dfe,tab:ablation_delamination_dfe,tab:ablation_fouling_dfe} show the performance of the classification network for Corrosion, Delamination, and Fouling defects respectively in different scenarios: \textbf{i)} without DFE, \textbf{ii)} with DFE but no regularization term in the training loss function and \textbf{iii)} with DFE and regularization term in the training loss function. In addition, \Cref{tab:ablation_corrosion_dfe_conf_mat,tab:ablation_delamination_dfe_conf_mat,tab:ablation_fouling_dfe_conf_mat} show the average confusion matrices values of the test sets at different cases of using DFE and regularization term for Corrosion, Delamination, and Fouling Defects, respectively.

The results indicate that the added Delamination Feature Extractor (DFE) can boost the performance of the network for the delamination type of defects that are most challenging to detect and classify. Moreover, it also boosts the overall performance of the network for other types of defects; namely corrosion and fouling. A justification for such a behavior can be that the General Feature Extractor (GFE) is now giving more focus to corrosion and fouling types of defects since the DFE is solely focusing on delamination defects. Moreover, since corrosion and fouling defects are generally identified by the texture of the defect unlike delamination which is typically identified by the boundaries, the GFE is focusing on extracting features related to the texture while the DFE is focusing on extracting features related to changes at the edges of a defect.

\begin{table}
    \centering
    \caption{Effectiveness of the Delamination Feature Extractor and Regularization on the detection of Corrosion Defects}
    \label{tab:ablation_corrosion_dfe}
    \begin{tabular}{r||c|c|c}
     & \multirow{2}{*}{without DFE} & \multicolumn{2}{c}{with DFE}\\
     & & no regularizer & with regularizer \\\hline
    Accuracy & 0.7636 
    & 0.7834 & \textbf{0.8009}\\
    B. Accuracy & 0.7854 & 0.7923 & \textbf{0.8153}\\
    F1-Score & 0.7872 & 0.8076 & \textbf{0.8319}\\\hline
    \end{tabular}
\end{table}

\begin{table}
    \centering
    \caption{Effectiveness of DFE and Regularization on the detection of Delamination Defects}
    \label{tab:ablation_delamination_dfe}
    \begin{tabular}{r||c|c|c}
     & \multirow{2}{*}{without DFE} & \multicolumn{2}{c}{with DFE}\\
     & & no regularizer & with regularizer \\\hline
    Accuracy & 0.6347  
    & 0.6789 & \textbf{0.7103}\\
    B. Accuracy & 0.5756 & 0.6250 & \textbf{0.6455}\\
    F1-Score & 0.5099 & 0.5808 & \textbf{0.6061}\\\hline
    \end{tabular}
\end{table}

\begin{table}
    \centering
    \caption{Effectiveness of the Delamination Feature Extractor and Regularization on the detection of Fouling Defects}
    \label{tab:ablation_fouling_dfe}
    \begin{tabular}{r||c|c|c}
     & \multirow{2}{*}{without DFE} & \multicolumn{2}{c}{with DFE}\\
     & & no regularizer & with regularizer \\\hline
    Accuracy & 0.7387 
    & 0.7697 & \textbf{0.7883}\\
    B. Accuracy & 0.7436 & 0.7866 & \textbf{0.7953}\\
    F1-Score & 0.7528 & 0.7759 & \textbf{0.8013}\\\hline
    \end{tabular}
\end{table}

% \newcommand{\confablation}[1]{%
% % \begin{figure}\centering
% \begin{subfigure}{\widthfig}\centering
% \includegraphics[width=\linewidth]{imgs/classification/ablations/#1/conf_mat/C.pdf}
% \caption{\scriptsize without DFE}
% \end{subfigure}
% \begin{subfigure}{\widthfig}\centering
% \includegraphics[width=\linewidth]{imgs/classification/ablations/#1/conf_mat/G.pdf}
% \caption{\scriptsize with DFE but no regularization}
% \end{subfigure}
% \begin{subfigure}{\widthfig}\centering
% \includegraphics[width=\linewidth]{imgs/classification/ablations/#1/conf_mat/B.pdf}
% \caption{\scriptsize with DFE and regularization}
% \end{subfigure}%
% \caption{Confusion matrices of {\it #1} defects for different cases of using DFE module and the regularization term.}
% \label{fig:ablation_#1_dfe_conf_mat}
% % \end{figure}%
% }
% \begin{figure*}
% \newcommand{\widthfig}{0.333\textwidth}
% \confablation{corrosion}
% \confablation{delamination}
% \confablation{fouling}
% \end{figure*}

%The min, mid and max values
\newcommand*{\MinNumber}{0.0}%
\newcommand*{\MidNumber}{0.5}%
\newcommand*{\MaxNumber}{1.0}%
%Apply the gradient macro
\newcommand{\ApplyGradient}[1]{%
        \ifdim #1 pt > \MidNumber pt
            \pgfmathsetmacro{\PercentColor}{max(min(100.0*(#1 - \MidNumber)/(\MaxNumber-\MidNumber),100.0),0.00)} %
            \hspace{-0.33em}\colorbox{DarkCyan!\PercentColor!Cyan}{$\mathbf{#1}$}
        \else
            \pgfmathsetmacro{\PercentColor}{max(min(100.0*(\MidNumber - #1)/(\MidNumber-\MinNumber),100.0),0.00)} %
            \hspace{-0.33em}\colorbox{LightCyan!\PercentColor!Cyan}{$\mathbf{#1}$}
        \fi
}

\newcolumntype{R}{>{\collectcell\ApplyGradient}c<{\endcollectcell}}
\renewcommand{\arraystretch}{0}
\setlength{\fboxsep}{3mm} % box size
\setlength{\tabcolsep}{0pt}

\newcommand{\confmatone}[5]{
\begin{subtable}{0.3\linewidth}
\caption{\scriptsize #1}
\vspace{-5pt}
\begin{tabular}{ccRR}
&  & \multicolumn{2}{c}{\scriptsize Prediction}\\
&  & \multicolumn{1}{c}{$(-)$} & \multicolumn{1}{c}{$(+)$}\\
\multirow{1}{*}{\parbox[t]{3mm}{\rotatebox[origin=c]{90}{\scriptsize Label}}} & \rotatebox[origin=c]{90}{$(-)$} & #2 & #3 \\
% \multirow{2}{*}{\parbox[b]{4mm}{\rotatebox[origin=c]{90}{ True Label}}} & \multicolumn{1}{c}{\parbox[t]{4mm}{\rotatebox[origin=c]{90}{$(-)$}}} & #2 & #3 \\
 & \rotatebox[origin=c]{90}{$(+)$} & #4 & #5 \\
\end{tabular}
\end{subtable}
}
\newcommand{\confmattwo}[5]{
\hspace{0.5em}
\begin{subtable}{0.25\linewidth}
\caption{\scriptsize #1}
\vspace{-5pt}
\begin{tabular}{RR}
\multicolumn{2}{c}{Prediction}\\
\multicolumn{1}{c}{$(-)$} & \multicolumn{1}{c}{$(+)$}\\
#2 & #3 \\
#4 & #5 \\
\end{tabular}
\end{subtable}
}

\begin{table}
\centering
\caption{Confusion matrices of {\it corrosion} defects for different cases of using DFE and regularization.}
\label{tab:ablation_corrosion_dfe_conf_mat}
\vspace{-5pt}
\confmatone{without DFE}{0.85}{0.15}{0.28}{0.72}
\confmattwo{DFE w/o regul.}{0.85}{0.15}{0.23}{0.73}
\confmattwo{DFE with regul.}{0.87}{0.13}{0.24}{0.76}
\vspace{-5pt}
\end{table}
\begin{table}
\centering
\caption{Confusion matrices of {\it delamination} defects for different cases of using DFE and regularization.}
\label{tab:ablation_delamination_dfe_conf_mat}
\vspace{-5pt}
\confmatone{without DFE}{0.68}{0.32}{0.53}{0.47}
\confmattwo{DFE w/o regul.}{0.72}{0.28}{0.47}{0.53}
\confmattwo{DFE with regul.}{0.75}{0.25}{0.46}{0.54}
\vspace{-5pt}
\end{table}
\begin{table}
\centering
\caption{Confusion matrices of {\it fouling} defects for different cases of using DFE and regularization.}
\label{tab:ablation_fouling_dfe_conf_mat}
\vspace{-5pt}
\confmatone{without DFE}{0.72}{0.28}{0.24}{0.76}
\confmattwo{DFE w/o regul.}{0.76}{0.24}{0.20}{0.80}
\confmattwo{DFE with regul.}{0.77}{0.23}{0.18}{0.82}
\vspace{-5pt}
\end{table}

% \begin{table}
% \caption{Confusion matrices of {\it corrosion} defects for different cases of using DFE module and the regularization term.}
% \vspace{-5pt}
% \begin{tabular}{ccRRcRRcRR}
% \rule{0pt}{15pt} & & \multicolumn{2}{c}{(a) without DFE} & & \multicolumn{2}{c}{(b) with DFE, no regul.} & & \multicolumn{2}{c}{(c) with DFE and regul.}\\
% \rule{0pt}{15pt} & & \multicolumn{2}{c}{Prediction}  & &  \multicolumn{2}{c}{Prediction} & & \multicolumn{2}{c}{Prediction}\\
% \rule{0pt}{15pt} & & \multicolumn{1}{c}{$(-)$} & \multicolumn{1}{c}{$(+)$} & & \multicolumn{1}{c}{$(-)$} & \multicolumn{1}{c}{$(+)$} & & \multicolumn{1}{c}{$(-)$} & \multicolumn{1}{c}{$(+)$}\\
% \multirow{2}{*}{\parbox[b]{4mm}{\rotatebox[origin=c]{90}{ True Label}}} & \multicolumn{1}{c}{\parbox[t]{4mm}{\rotatebox[origin=c]{90}{$(-)$}}} & 0.85 & 0.15 & & 0.85 & 0.15 & & 0.87 & 0.13\\
%  & \multicolumn{1}{c}{\parbox[t]{4mm}{\rotatebox[origin=c]{90}{$(+)$}}} & 0.28 & 0.72 & & 0.23 & 0.73 & & 0.24 & 0.76 \\
% \end{tabular}
% \end{table}

\renewcommand{\arraystretch}{1}
\setlength{\tabcolsep}{3pt}

\subsubsection{\textbf{Effectiveness of the STN modules in the classification network}}\label{subsubsec:ablation_stn} 
We also perform another ablation study to confirm the viability of the STN module. Although we have noticed that the STN module is not as important as the novel DFE module and regularization in terms of increasing the accuracy and balanced accuracy, it was able to give a consistent increase in those metrics for all types of defects. The reason for that better classification could be the limited training set which gives more importance to the STN module. To elaborate, STN module's sole purpose is to learn an affine transformation performed over the input patch so that it can be classified accurately down the line. Its importance magnifies with the smaller size of the training set.

\begin{figure*}
\newcommand{\widthfig}{0.333\textwidth}
\begin{subfigure}{\widthfig}\centering
\includegraphics[width=\linewidth]{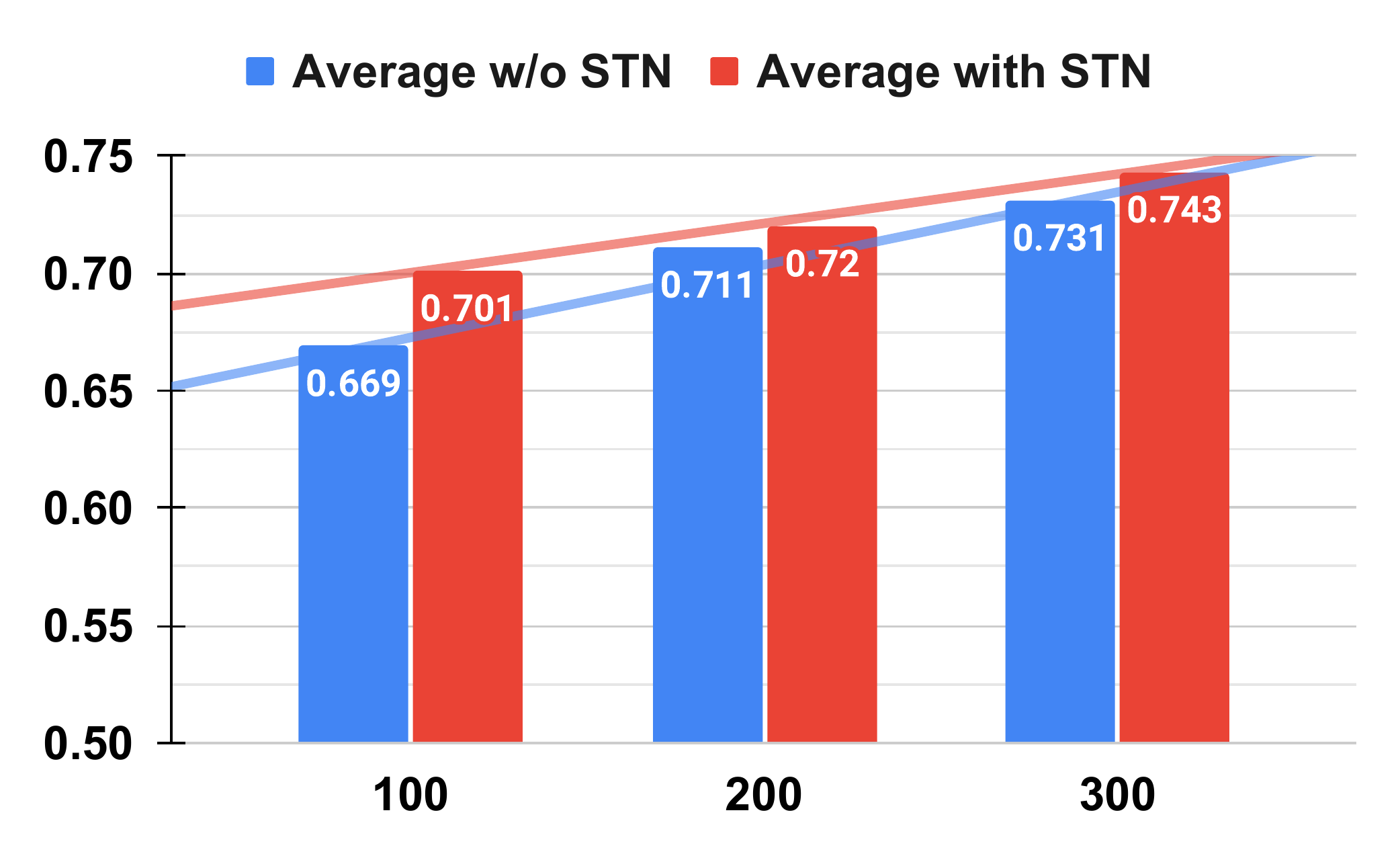}
\caption{\scriptsize Accuracy}
\end{subfigure}
\begin{subfigure}{\widthfig}\centering
\includegraphics[width=\linewidth]{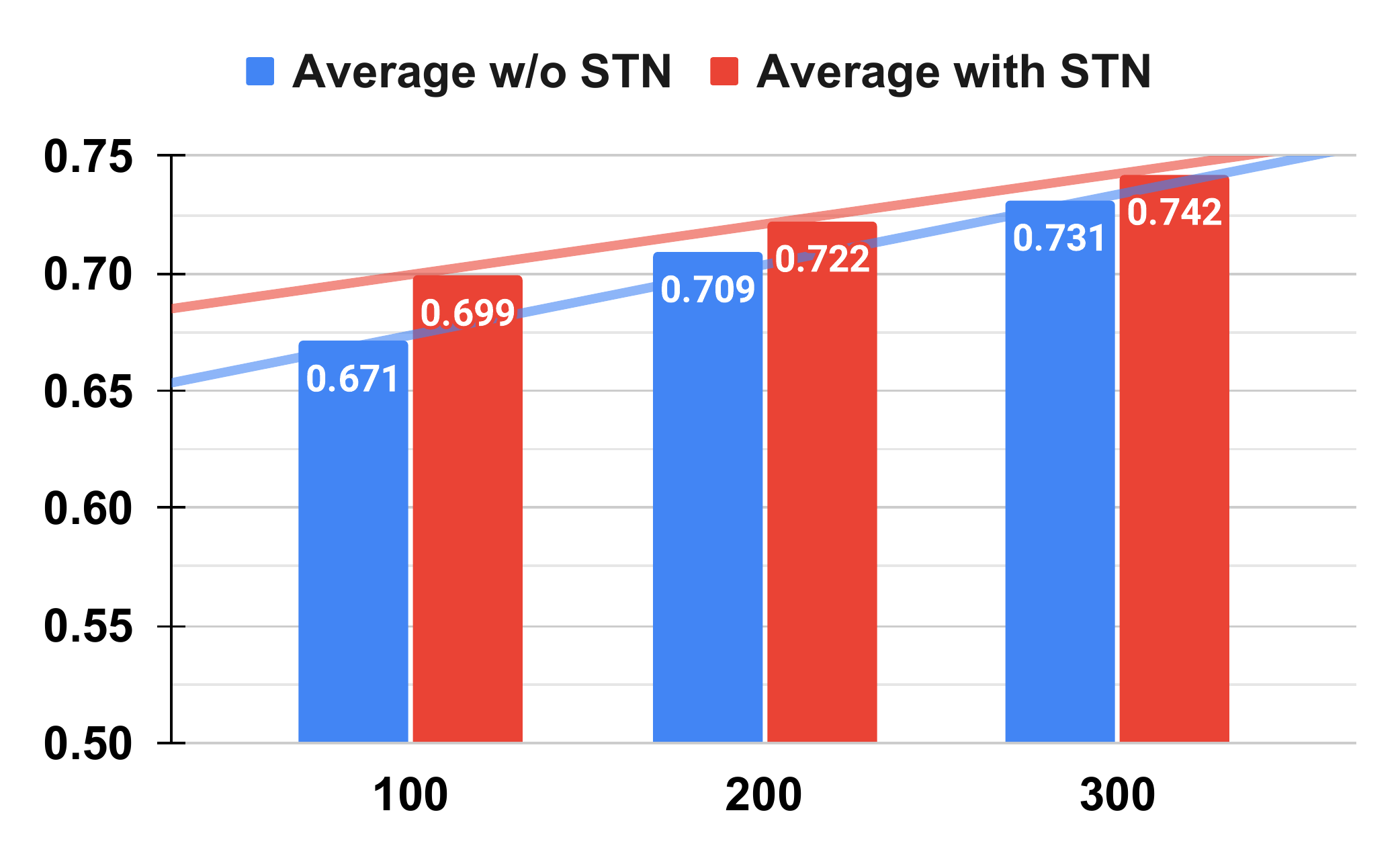}
\caption{\scriptsize Balanced Accuracy}
\end{subfigure}
\begin{subfigure}{\widthfig}\centering
\includegraphics[width=\linewidth]{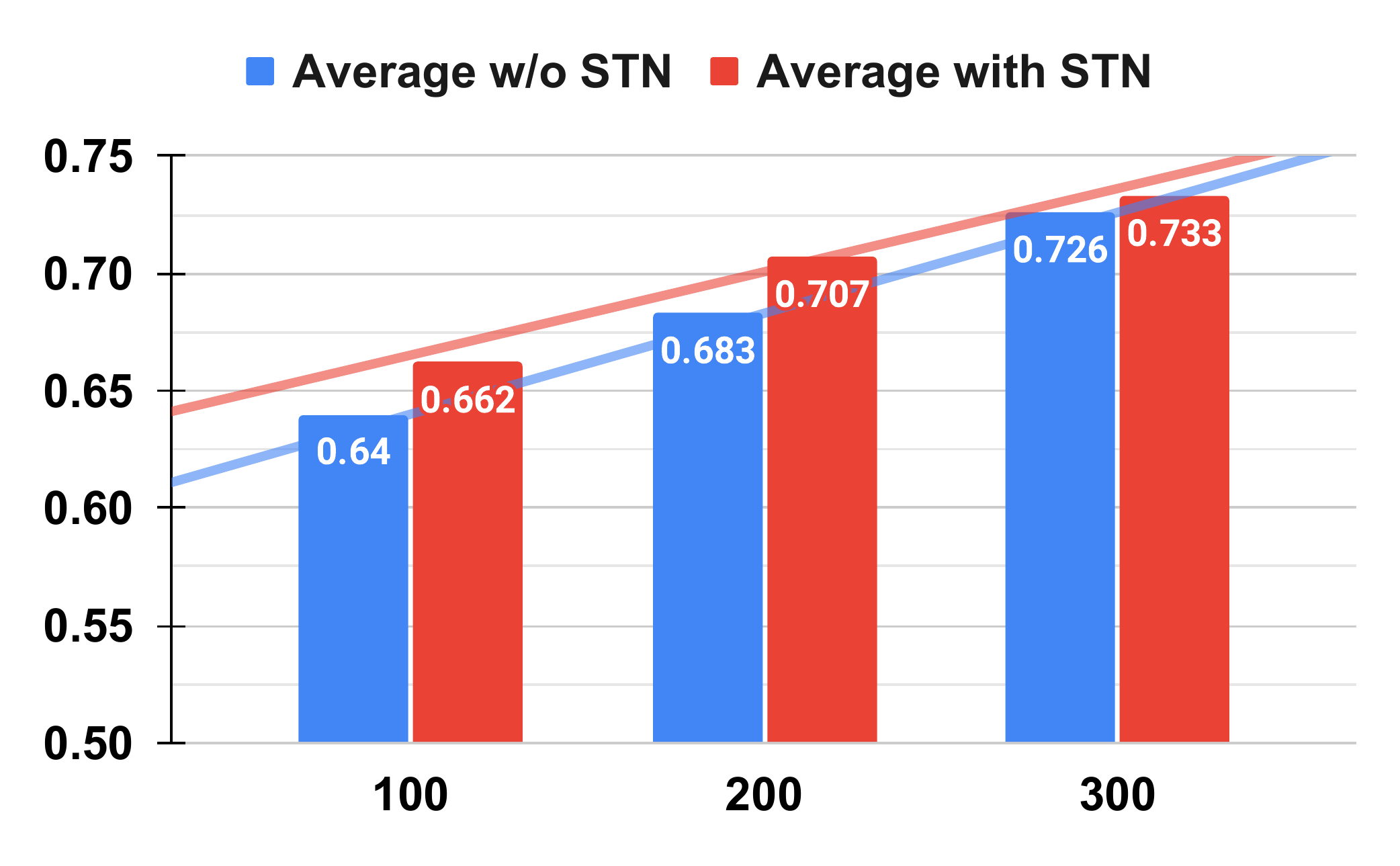}
\caption{\scriptsize F1-Score}
\end{subfigure}%
\caption{Performance evaluation of the classification network with and without STN module at different sizes of the training dataset. The shown results are the average over the three defect types.}
\label{fig:ablation_stn}
\end{figure*}

To validate this conclusion, we have performed a limited training case study, where we have only used 100, 200, and 300 sets of images for training and used 30 images as a test set. Then, we measured the accuracy, balanced accuracy, and F1-score for each set. The results of each test set are depicted in \Cref{fig:ablation_stn}, where the average is taken over the three types of defects. As can be shown from the figure, albeit the STN module boosts the performance, but its effect is much more significant by reducing the training dataset size. Because The network (without STN module) cannot generalize well with a smaller training set and it is much prone to overfitting. Therefore, the STN module has a similar-to regularization effect in the case of a smaller training set. As shown in \Cref{fig:ablation_stn}, the STN module allows the network to have much more graceful decay in performance when the size of the training set is reduced by a factor of 3.

\subsubsection{\textbf{Utilization of the section segmentation output in the classification process}} 
\begin{figure*}
\begin{subfigure}{\linewidth}
\includegraphics[width=\linewidth]{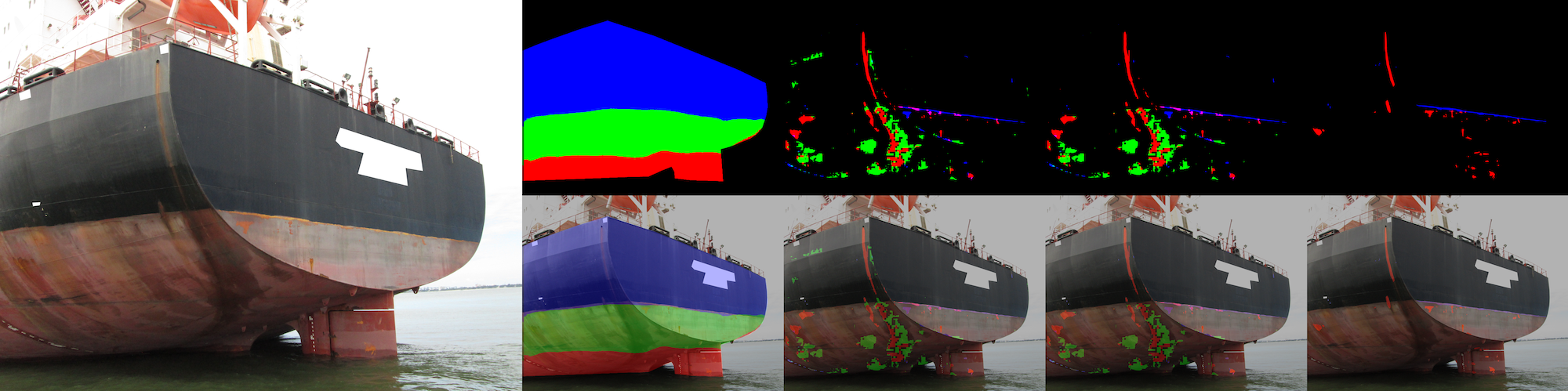}
\caption{}
\label{fig:POC 1}
\end{subfigure}
\begin{subfigure}{\linewidth}
\includegraphics[width=\linewidth]{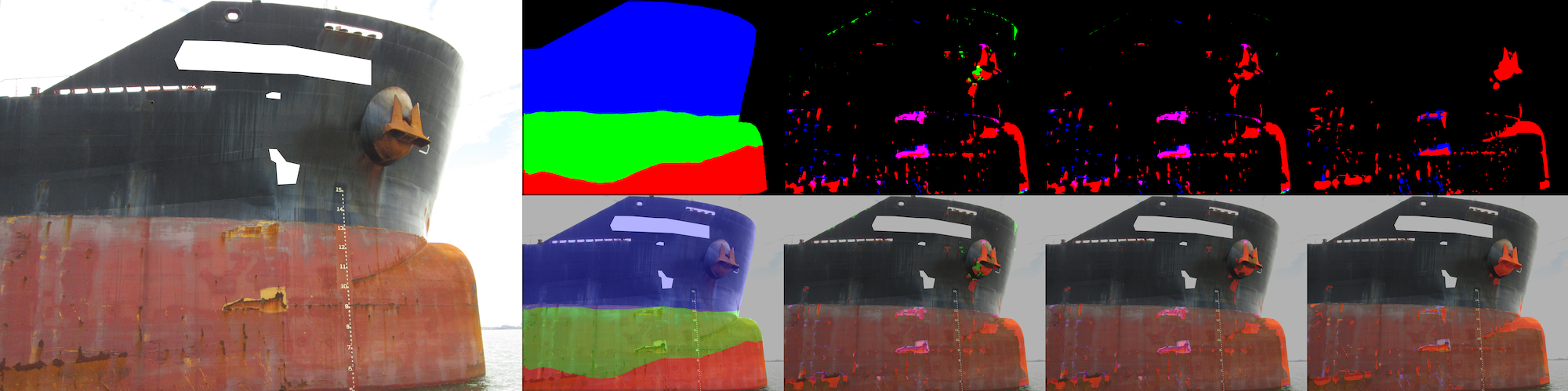}
\caption{}
\label{fig:POC 3}
\end{subfigure}
\begin{subfigure}{\linewidth}
\includegraphics[width=\linewidth]{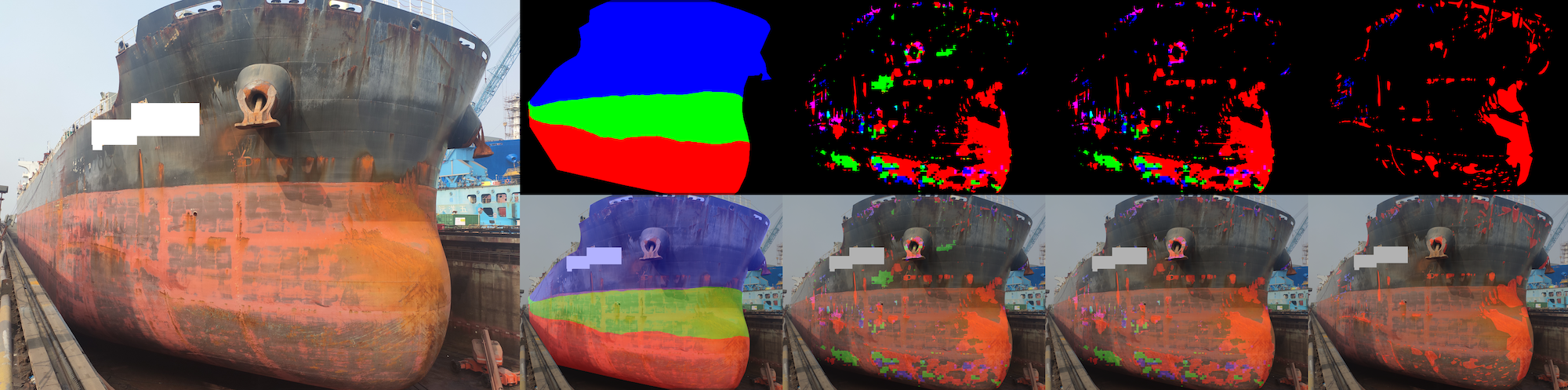}
\caption{}
\label{fig:POC 11}
\end{subfigure}
\begin{subfigure}{\linewidth}
\includegraphics[width=\linewidth]{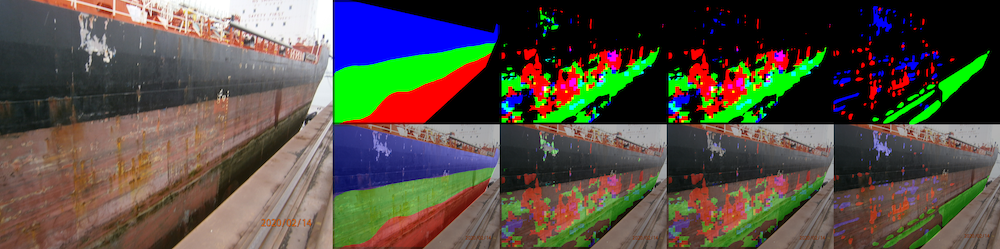}
\caption{}
\label{fig:POC 28}
\end{subfigure}
\caption{Four examples of the test set. From left to right, i) input image, ii) section segmentation results, iii) defect classification without employing section segmentation, iv) defect classification employing section segmentation and v) ground-truth labels. The first row of each sample shows the results/labels and the second row shows the overlaid results/labels over the input image.}
\label{fig:example_fouling_ablation}
\end{figure*}

We have noticed that the distribution of defects varies depending on the location; whether at TS, BT, or VS. Prominently, Fouling defects never happen in the the TS section of the ship because TS is not under water where the fouling organisms are. 
% We have found that the average percentage of the fouling defected area is $0.42 \%$ in TS section and $20.16 \%$ and $42.82 \%$ in BT and VS sections, respectively. Moreover, only 2 samples of the 60 images have fouling defects in the TS section that exceed $1\%$ of the area. This confirms the validity of neglecting the fouling defects from the TS section. 
% \Cref{fig:fouling_ts_60images} shows the percentage of fouling defected area in the TS section of those 60 images. It is clear that disregarding fouling defects in the TS section of ships should lead to better detection results. 
At post-processing, we remove all fouling from the TS section.
We show four different examples of a test set in \Cref{fig:example_fouling_ablation}. Notably, the network (without employing section segmentation results) detects incorrect fouling defects in the top section of the ship (TS), while it is typically rare to occur. Therefore, the results of the network after employing the section segmentation results are much more consistent and close to the ground truth labels. 
% It is noteworthy that in some cases the results of \clsName are even more faithful than the provided ground-truths (according to experts from PPG Industries).

% \begin{figure}
%     \centering
%     \includegraphics[width=\linewidth]{imgs/classification/ablations/part_seg/part_seg_inclusion.pdf}
%     \caption{Percentage of fouling defected area in TS section of 60 test vessel images.}
%     \label{fig:fouling_ts_60images}
% \end{figure}

\subsection{Results}\label{subsec:result}
\subsubsection{Ship section segmentation}
We compare our model with two baselines, UNet and HorizonNet. The result is evaluated as mean IoU of TS, BT, and VS over 35 test samples, as shown in Table~\ref{tab:partiou}. From the mean IoUs, we know our method is consistently better than the two baselines. Note that the prediction from our model is two 1D vectors and they have to be combined with the whole ship segmentation to get the 2D segmentation map.
\begin{table}
    \centering
    \caption{Mean IoU of TS/BT/VS}
    \begin{tabular}{l|c|c|c}
        \hline
        Method & TS & BT & VS \\
        \hline
        UNet~\cite{ronneberger2015u} & 0.7459 & 0.4669 & 0.5360 \\
        \hline
        HorizonNet~\cite{sun2019horizonnet} & 0.7827 & 0.5717 & 0.6800 \\
        \hline
        Ours & 0.8542 & 0.6641 & 0.7535 \\
        \hline
    \end{tabular}
    \label{tab:partiou}
\end{table}

\subsubsection{Defect segmentation}
We also compare the result in defect segmentation with one baseline UNet. Since we use UNet architectures for both teacher and student models, the baseline here is equal to the teacher model. For the comparison, we selected 60 vessel images and the result is evaluated as three metrics: mean IoU, precision, and recall, as shown in Table~\ref{tab:defectseg}. Because of the improvement of labels from the teacher model, our student model achieves a better recall rate and IoU. The precision does not improve much because the inclusion of pseudo labels inevitably introduced noise that makes the student tends to predict more defects than the teacher. This leads to the rise in recall rate at the sacrifice of the precision score.
\begin{table}
    \centering
    \caption{Mean IoU, precision, and recall rates}
    \begin{tabular}{l|c|c|c}
        \hline
        Method &  IoU & Precision & Recall \\
        \hline
        UNet~\cite{ronneberger2015u} & 0.4924 & 0.6373 & 0.6523 \\
        \hline
        Ours & 0.5205 & 0.6314 & 0.7725 \\
        \hline
    \end{tabular}
    \label{tab:defectseg}
\end{table}

\subsubsection{Defect classification}
To the best of our knowledge, there is no current published method that can detect defects of marine vessels. Therefore, it is not possible to compare the performance of our network with another state-of-the-art method. However, to have a justified and fair evaluation of our classification algorithm, we compare \clsName with both ResNet \cite{he2016deep} and DenseNet \cite{huang2017densely} based implementation without the novel DFE additional encoder nor the STN module. Moreover, to ensure the reliability of the network in a real-life scenario, we compare the performance of our network to human experts in the detection of defects in marine vessels for 60 non-labeled images of marine vessels.

\Cref{tab:results_corrosion_conf_mat,tab:results_delamination_conf_mat,tab:results_fouling_conf_mat} show the confusion matrices of different types of defects for our network in comparison to DenseNet- and ResNet-based architectures. The results are calculated over 50 labeled test images. It is clear that our network performs better than ResNet and DenseNet based networks; mainly due to the DFE and the STN module as has been discussed in \Cref{subsubsec:ablation_dfe,subsubsec:ablation_stn}.

\begin{figure*}
\centering
\begin{subfigure}{1.0\textwidth}\centering
\includegraphics[width=\linewidth]{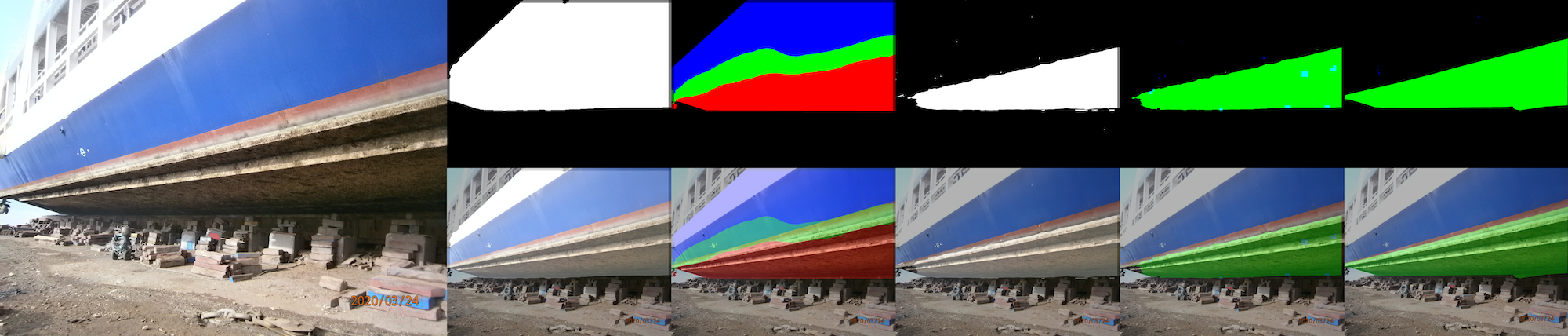}
% \caption{\scriptsize example 1}
\end{subfigure}
\begin{subfigure}{1.0\textwidth}\centering
\includegraphics[width=\linewidth]{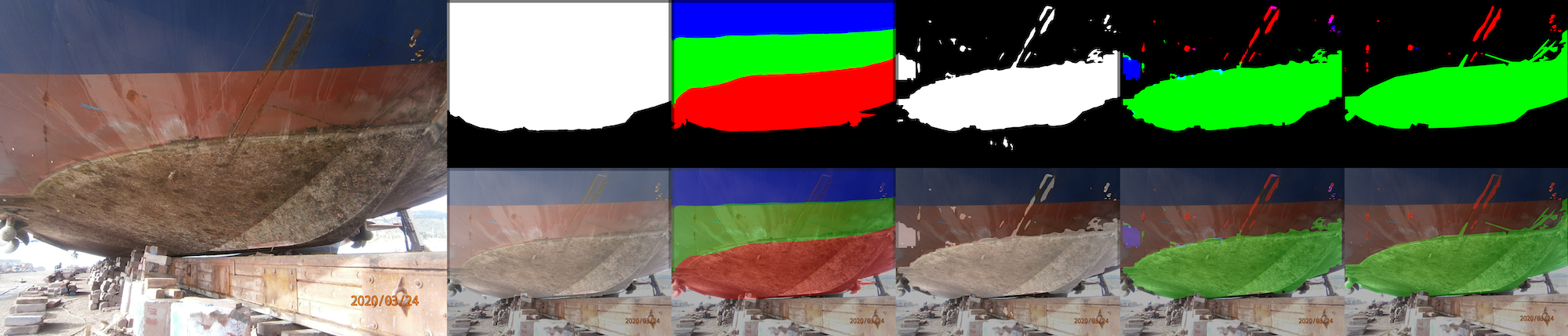}
% \caption{\scriptsize example 2}
\end{subfigure}
\begin{subfigure}{1.0\textwidth}\centering
\includegraphics[width=\linewidth]{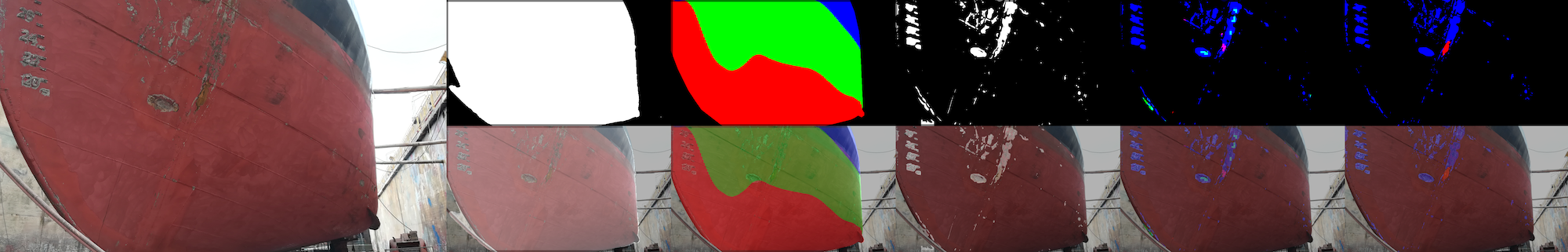}
% \caption{\scriptsize example 2}
\end{subfigure}
\begin{subfigure}{1.0\textwidth}\centering
\includegraphics[width=\linewidth]{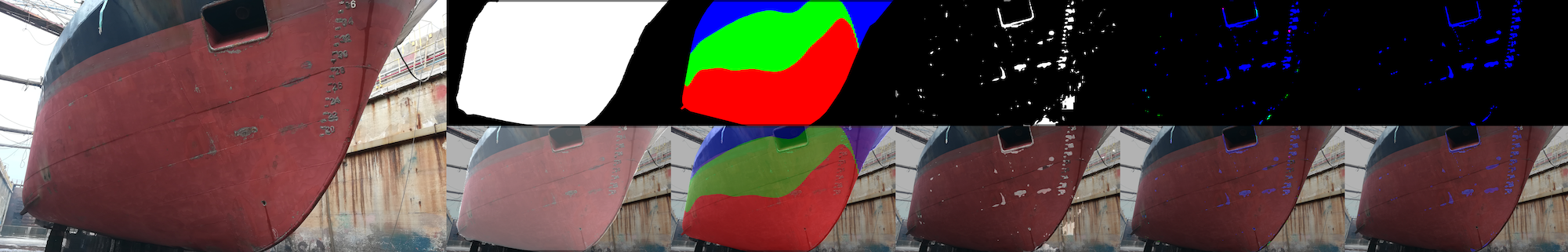}
% \caption{\scriptsize example 2}
\end{subfigure}
\caption{Four examples that the algorithm is doing the same as the provided labeled data.}
\label{fig:extra_cls_examples_same}
\end{figure*}

The second experiment that we have performed is to test the performance of \clsName compared to human experts. With the help of experts in the field from PPG, we were able to obtain the percentages of defects in each section of 60 vessel images. These results were calculated by 6 human experts in the analysis of vessel defects. 
% It was cumbersome for the experts to label these images, thus they have only provided the percentages of defects in each section of each vessel in the 60 non-labeled images.
\Cref{tab:results_percentages_POC} shows the predicted percentage of defects of different methods in comparison with the human experts. It is noteworthy that not all of the images have all sections of the vessel. Therefore, the average of defects at a section is taken considering only images that contain this section only. For instance, an image may not contain a TS section. Thus, we will not include it in the statistical calculations of the percentage of the defected area of the TS section. This leads to an effectively lower number of images for the statistical analysis and thus less confidence in the results. However, it is not that significant, as the number of images that includes a specific section is more than 40, which is more than enough for high confidence in the analysis.
\Cref{fig:extra_cls_examples_same} shows four examples from the test set, where \clsName was successful to obtain results that match the provided labeled data. 
% In \Cref{fig:extra_cls_examples_better}, we show three examples from the test set where it seems that \clsName results are not very consistent with the provided labeled data. However, after careful inspection, the results of \clsName are better than the provided labeled data itself. We have also shown the results to experts in the field at PPG Industries and they have confirmed the validity of our claims.

\begin{table*}
\centering
\caption{Section wise average defect percentage across 60 vessel images. Note that multiple experts rate each image, hence we first compute the average over the 60 images of each inspector, then we compute the mean and standard deviation over the average ratings of the inspectors. Experts values are written as mean $\mu$ (std. dev. $\sigma$).}
\label{tab:results_percentages_POC}
\resizebox{\textwidth}{!}{%
\begin{tabular}{|r||gcg|cgc|gcg|}\hline
    Defect & \multicolumn{3}{c|}{\textbf{Corrosion}} & \multicolumn{3}{c|}{\textbf{Delamination}} & \multicolumn{3}{c|}{\textbf{Fouling}}\\\hline
    section & TS & BT & VS & TS & BT & VS & TS & BT & VS \\\Xhline{4\arrayrulewidth}
    Experts & 
    11.08 (3.34) & 20.36 (5.09) & 8.90 (4.98) & 
    3.84 (1.09) & 13.67 (3.90) & 25.83 (8.41) &
    0.42 (0.01) & 35.12 (4.80) & 66.03 (9.17)
    \\\hline
    ResNet \cite{he2016deep} & 
    6.08 & 12.08 & 5.98 & 
    2.18 & 8.07 & 18.04 &
    3.74 & 31.08 & 54.30
    \\\hline
    DenseNet \cite{huang2017densely} & 
    8.81 & 15.30 & 5.72 & 
    3.14 & 7.94 & 19.46 &
    5.31 & 29.50 & 56.72 
    \\\Xhline{4\arrayrulewidth}
    % \rowcolor{LightCyan}
    \clsName & 
    9.84 & 18.52 & 8.01 &
    5.20 & 10.12 & 22.74 &
    0 & 38.17 & 61.82
    \\\hline
\end{tabular}%
}
\end{table*}

% \newcommand{\confres}[1]{%
% % \begin{figure}\centering
% \begin{subfigure}{\widthfig}\centering
% \includegraphics[width=\linewidth]{imgs/classification/results/#1/conf_mat/ResNet.pdf}
% \caption{\scriptsize ResNet-based architecture}
% \end{subfigure}
% \begin{subfigure}{\widthfig}\centering
% \includegraphics[width=\linewidth]{imgs/classification/results/#1/conf_mat/DenseNet.pdf}
% \caption{\scriptsize DenseNet-based architecture}
% \end{subfigure}
% \begin{subfigure}{\widthfig}\centering
% \includegraphics[width=\linewidth]{imgs/classification/ablations/#1/conf_mat/B.pdf}
% \caption{\scriptsize \clsName}
% \end{subfigure}%
% \caption{Confusion matrices of {\it #1} defects for different architectures.}
% \label{fig:results_#1_conf_mat}
% % \end{figure}%
% }
% \begin{figure*}
% \newcommand{\widthfig}{0.333\textwidth}
% \confres{corrosion}
% \confres{delamination}
% \confres{fouling}
% \end{figure*}

\renewcommand{\arraystretch}{0}
\setlength{\fboxsep}{3mm} % box size
\setlength{\tabcolsep}{0pt}

\begin{table}
\centering
\caption{Confusion matrices of {\it corrosion} defects for different architectures.}
\label{tab:results_corrosion_conf_mat}
\vspace{-5pt}
\confmatone{ResNet-based}{0.82}{0.18}{0.29}{0.71}
\confmattwo{DenseNet-based}{0.84}{0.16}{0.30}{0.70}
\confmattwo{\clsName}{0.87}{0.13}{0.24}{0.76}

\caption{Confusion matrices of {\it delamination} defects for different architectures.}
\label{tab:results_delamination_conf_mat}
\vspace{-5pt}
\confmatone{ResNet-based}{0.67}{0.33}{0.54}{0.46}
\confmattwo{DenseNet-based}{0.65}{0.35}{0.58}{0.42}
\confmattwo{\clsName}{0.75}{0.25}{0.46}{0.54}

\caption{Confusion matrices of {\it fouling} defects for different architectures.}
\label{tab:results_fouling_conf_mat}
\vspace{-5pt}
\confmatone{ResNet-based}{0.71}{0.29}{0.26}{0.74}
\confmattwo{DenseNet-based}{0.72}{0.28}{0.26}{0.74}
\confmattwo{\clsName}{0.77}{0.23}{0.18}{0.82}
\end{table}

\renewcommand{\arraystretch}{1}
\setlength{\tabcolsep}{3pt}

% \begin{figure*}
% \centering
% \begin{subfigure}{0.9\textwidth}\centering
% \includegraphics[width=\linewidth]{imgs/classification/results/better_than_gt/POC 1.png}
% % \caption{\scriptsize example 1}
% \end{subfigure}
% \begin{subfigure}{0.9\textwidth}\centering
% \includegraphics[width=\linewidth]{imgs/classification/results/better_than_gt/POC 2.png}
% % \caption{\scriptsize example 2}
% \end{subfigure}
% \begin{subfigure}{0.9\textwidth}\centering
% \includegraphics[width=\linewidth]{imgs/classification/results/better_than_gt/POC 17.png}
% % \caption{\scriptsize example 2}
% \end{subfigure}
% \caption{Three examples that the algorithm is doing much better than the provided labeled data.}
% \label{fig:extra_cls_examples_better}
% \end{figure*}

\subsubsection{Discussion about multi-label vs. multi-class architecture} 
We have adopted a multi-label approach in the classification process of defects. However, another possible approach is to use a multi-class approach; which only detects a single class (the most prominent class) in a patch. The multi-class approach answers one question: What type of defect in the patch, corrosion, delamination, or fouling? The multi-label approach attempts to answer three questions instead: (1) Does the patch have a corrosion defect? (2) Does the patch have a delamination defect? and (3) Does the patch have a fouling defect? 

We show here a comparison of both approaches. In conclusion, using a multi-label approach is better than a multi-class one. There are two reasons for that. First, the percentage of patches that have overlapping defects (at least two defects) cannot be neglected. Second, the defect segmentation algorithm may falsely detect defects. Multi-label classification can further clean some of those erroneously detected defects by answering its three questions by No.

For the comparison, we have selected 20 images as a test set. To elaborate on the importance of the multi-label approach, those selected images have overlapping defects. Thus, many of the generated patches of those images have multiple labels. \Cref{tab:multi_class_multi_label} shows the accuracy, the balanced accuracy, and the F1-score in both cases. The multi-label approach is better for the detection and classification of defects in marine vessels for the aforementioned reasons. 
% We also show in \Cref{fig:multi_class_multi_label} a couple of examples to show how both approaches label defects.

\begin{table}
    \caption{Numerical results of multi-class approach vs multi-label approach}
    \label{tab:multi_class_multi_label}
    \centering
    \begin{tabular}{r|c|c|c|}
                & Accuracy & B. Accuracy & F1-Score \\\Xhline{4\arrayrulewidth}
    Multi-Class & 0.62 & 0.54 & 0.63 \\\hline
    \rowcolor{LightCyan}
    Multi-Label & \textbf{0.78} & \textbf{0.76} & \textbf{0.72} \\\hline
    \end{tabular}
\end{table}

% \begin{figure}
% \begin{subfigure}{0.49\textwidth}\centering
% \includegraphics[width=\linewidth]{imgs/classification/ablations/multi-class/ex1.png}
% \caption{\scriptsize }
% \end{subfigure}
% \begin{subfigure}{0.49\textwidth}\centering
% \includegraphics[width=\linewidth]{imgs/classification/ablations/multi-class/ex2.png}
% \caption{\scriptsize }
% \end{subfigure}
% \caption{Two examples to highlight the difference between multi-class and multi-label approaches}
% \label{fig:multi_class_multi_label}
% \end{figure}

\section{Conclusions and Future Work}\label{sec:conclusion}
In this paper, we built the first dataset for surface defect detection of marine vessels and proposed a multi-stage deep learning framework to evaluate the percentage of three types of defects over three sections of a ship. The prediction of sections (TS, BT, VS) of a ship is converted from 2D masks to 1D boundaries so that less training data is needed and the result is more consistent. We utilized a teacher-student training scheme for defect segmentation to tackle the problem of coarse and incomplete labels. A teacher model is firstly trained on labeled data and used to generate pseudo labels. The generated labels are then combined with original labels to train a student model to achieve better performance. For defect classification, we proposed a multi-label classification network that includes two detection heads for challenging defect (delamination) detection and STNs to estimate affine transformation for input patches before feature extraction such that patches are warped for better alignment. Our individual modules achieved better performance than state-of-the-art methods and comparable performance to human inspectors.

Aspects of our approach can be further improved. An end-to-end framework in place of the defect segmentation and classification modules may be desired and possibly reduce inference cost. The domain knowledge that fouling usually happens at the lower part of a ship can be utilized in other ways. Currently, we make use of it by post-processing to remove fouling from TS. Future work may consider estimating the relative position of fouling in the ship and enforce it in loss function or remove it before final prediction.

\newpage

\section*{Acknowledgment}

The authors would like to thank Phillip C. Yu, Stephen Aro, Melinda Dent (Shearer), Theresa Walker, Periklis Labrinides, Paulo Cardoso, Michael MacDonald, Joseph Agustin, Daniel Ihada, Dan Robbins, Rene Blom, Marta Lourenco, and others of the PPG Industries, Inc. for the valuable discussions and for providing the carefully annotated marine vessels dataset. We  also appreciate the input and the feasibility experiments conducted by Dolzodmaa Davaasuren, Tiantong Guo, and Zhuomin Zhang of Penn State.

\bibliographystyle{IEEEtran}
\bibliography{ref}

\begin{IEEEbiography}[{\includegraphics[width=1.1in,clip,keepaspectratio]{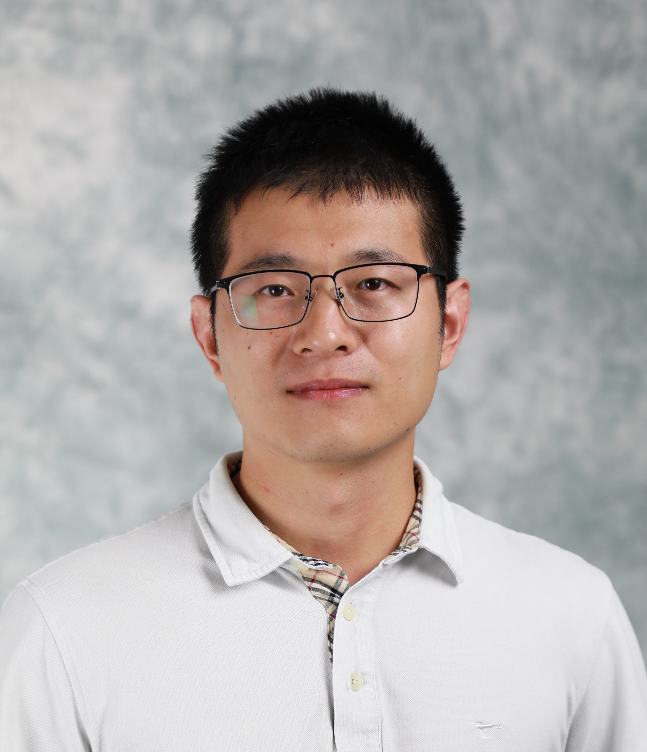}}]
{Li Yu} received his MS degree in Informatics from the School of Information Sciences and Technology at The Pennsylvania State University. He received his bachelor's degree with double major in chemistry and mathematics from Peking University and the MS degree in chemistry from Emory University. His research interests lie in the field of computer vision and image processing.
\end{IEEEbiography}

\begin{IEEEbiography}[{\includegraphics[width=1.1in,clip,keepaspectratio]{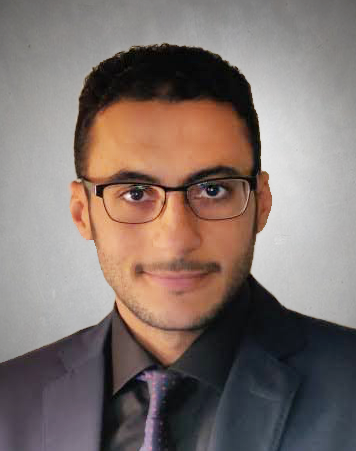}}]
{Kareem Metwaly} is a PhD candidate at the School of Electrical Engineering and Computer Science, The Pennsylvania State University. He obtained both his bachelor's degree in Electrical Engineering and MS degree in Engineering Mathematics from Alexandria University, Egypt. His research interests lie in the intersection of machine learning, specifically deep learning, with computer vision and image processing. He worked previously in Brightskies Technologies, Alexandria, Egypt for two years, where he worked on implementing mathematically and computationally high-performance algorithms. He is currently doing an internship at Toyota Motor North America, where he is using machine learning strategies and approaches to solve some computer vision tasks for vehicles. More information about him is at \url{http://personal.psu.edu/kmm1122}
\end{IEEEbiography}

\begin{IEEEbiography}[{\includegraphics[width=1.1in,clip,keepaspectratio,trim=2 12 0 0]{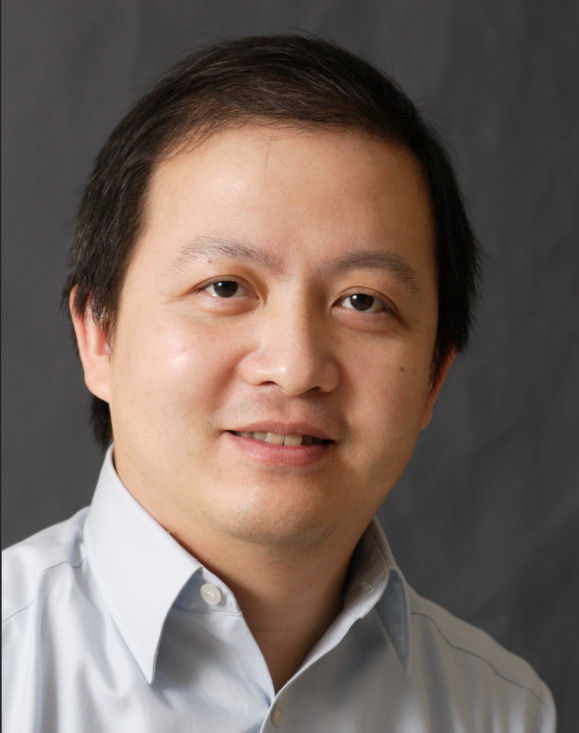}}]
{James Z. Wang}
is a Professor of Information Sciences and Technology at The Pennsylvania State University. He received the bachelor's degree in mathematics {\it summa cum laude} from the University of Minnesota, and the MS degree in mathematics, the MS degree in computer science, and the PhD degree in medical information sciences, all from Stanford University. His research interests include image analysis, image modeling, image retrieval, and their applications. He was a visiting professor at the Robotics Institute at Carnegie Mellon University (2007-2008), a lead special section guest editor of the IEEE Transactions on Pattern Analysis and Machine Intelligence (2008), and a program manager at the Office of the Director of the National Science Foundation (2011-2012). He was a recipient of a National Science Foundation Career award (2004) and Amazon Research Awards (2018, 2019).
\end{IEEEbiography}

\begin{IEEEbiography}[{\includegraphics[width=1.1in,height=1.3in,clip,keepaspectratio]{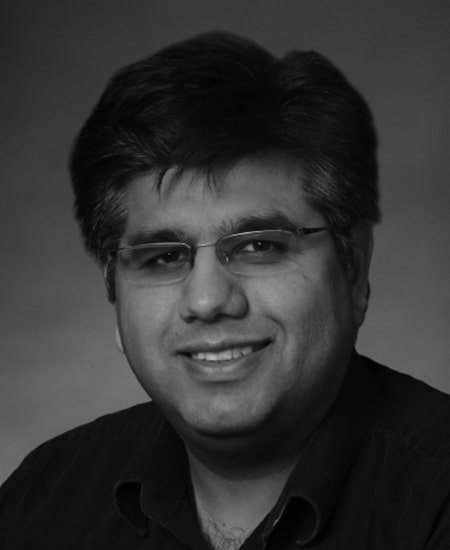}}]
{Vishal Monga}
received his Ph.D. degree in electrical engineering from the University of Texas at Austin. Currently, he is a professor in the School of Electrical Engineering and is a member of the electrical engineering and computer science faculty at Pennsylvania State University, University Park, Pennsylvania, 16802, USA. He is an elected member of the IEEE Image Video and Multidimensional Signal Processing Technical Committee and a senior area editor of IEEE Signal Processing Letters, and he has served on the editorial boards of IEEE Transactions on Image Processing, IEEE Signal Processing Letters, and IEEE Transactions on Circuits and Systems for Video Technology. He is a recipient of the U.S. National Science Foundation CAREER Award and a 2016 Joel and Ruth Spira Teaching Excellence Award. His research interests include optimization-based methods with applications in signal and image processing, learning, and computer vision. He is a Senior Member of IEEE.
\end{IEEEbiography}

\end{document}